\documentclass[10pt,twocolumn,letterpaper]{article}

\usepackage[pagenumbers]{cvpr} 

\usepackage[accsupp]{axessibility}
\usepackage{times}
\usepackage{epsfig}
\usepackage{graphicx}
\usepackage{amsmath}
\usepackage{amssymb}
\usepackage{enumitem}
\usepackage{booktabs}
\usepackage[font=small,skip=0pt]{caption}
\usepackage[dvipsnames, table]{xcolor}
\usepackage{multirow}
\usepackage{pifont}
\newcommand{\cmark}{\ding{51}}
\newcommand{\xmark}{\ding{55}}
\definecolor{light-gray}{gray}{0.85}

\usepackage[pagebackref,breaklinks,colorlinks]{hyperref}

\usepackage[capitalize]{cleveref}
\crefname{section}{Sec.}{Secs.}
\Crefname{section}{Section}{Sections}
\Crefname{table}{Table}{Tables}
\crefname{table}{Tab.}{Tabs.}

\def\cvprPaperID{9423} 
\def\confName{CVPR}
\def\confYear{2022}

\begin{document}

\title{Leveraging Real Talking Faces via Self-Supervision for Robust Forgery Detection}

\author{\stepcounter{footnote}Alexandros Haliassos\textsuperscript{1,}\thanks{Corresponding author.}
\and
Rodrigo Mira\textsuperscript{1}
\and
Stavros Petridis\textsuperscript{1,2}
\and
Maja Pantic\textsuperscript{1,2}
\and
\textsuperscript{1}Imperial College London
\and
\textsuperscript{2}Meta AI
\and
{\tt\small \{alexandros.haliassos14,rs2517,stavros.petridis04,m.pantic\}@imperial.ac.uk}
}
\maketitle

\begin{abstract}
One of the most pressing challenges for the detection of face-manipulated videos is generalising to forgery methods not seen during training while remaining effective under common corruptions such as compression. In this paper, we examine whether we can tackle this issue by harnessing videos of \textit{real} talking faces, which contain rich information on natural facial appearance and behaviour and are readily available in large quantities online. Our method, termed RealForensics, consists of two stages. First, we exploit the natural correspondence between the visual and auditory modalities in real videos to learn, in a self-supervised cross-modal manner, temporally dense video representations that capture factors such as facial movements, expression, and identity. Second, we use these learned representations as targets to be predicted by our forgery detector along with the usual binary forgery classification task; this encourages it to base its real/fake decision on said factors. We show that our method achieves state-of-the-art performance on cross-manipulation generalisation and robustness experiments, and examine the factors that contribute to its performance. Our results suggest that leveraging natural and unlabelled videos is a promising direction for the development of more robust face forgery detectors. Code is available.\footnote{ {\url{https://github.com/ahaliassos/RealForensics}}}
 
\end{abstract}

\section{Introduction}
\label{sec:intro}

\begin{figure}[t]
\begin{center}
  \includegraphics[width=\linewidth]{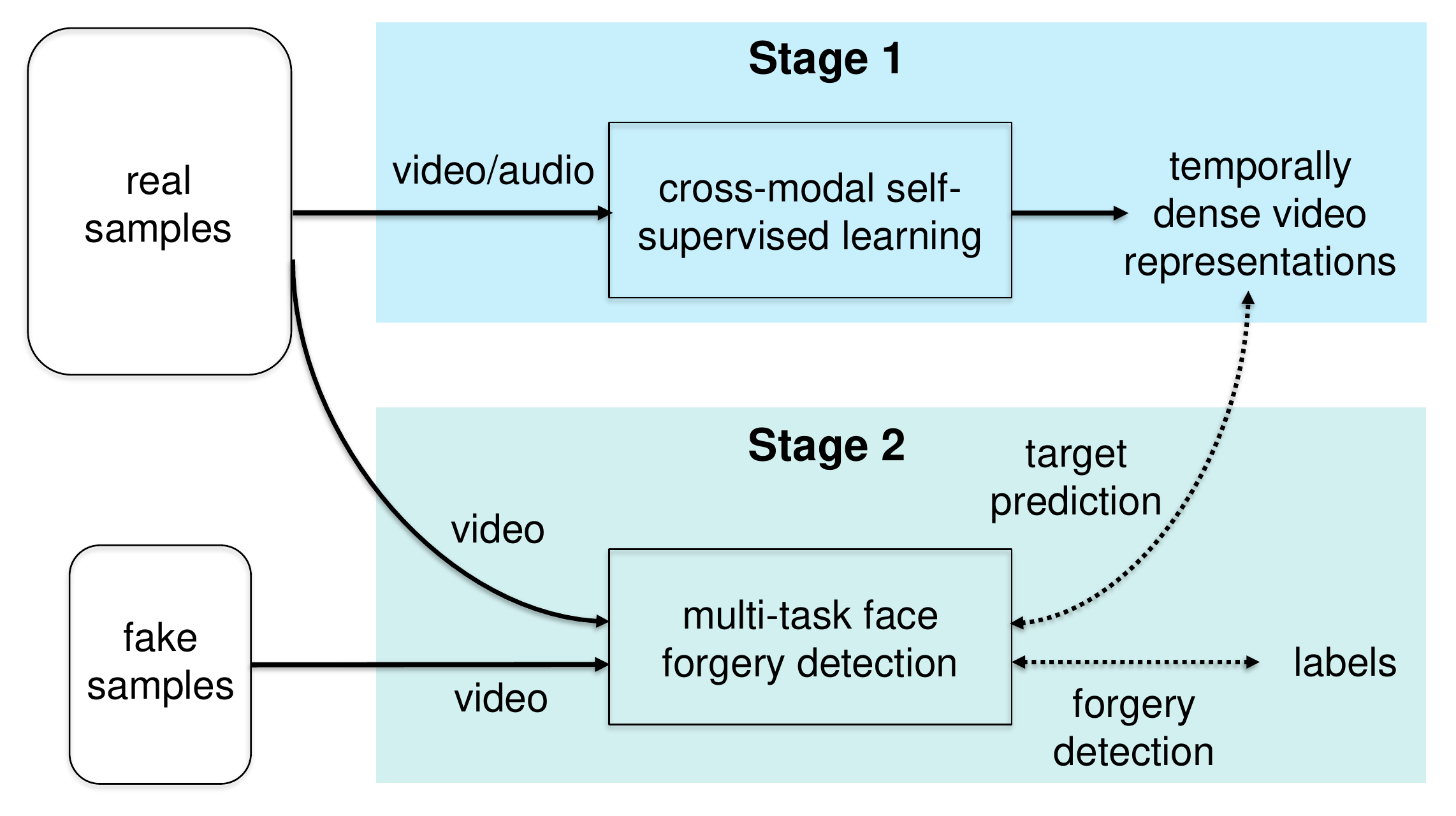}
\end{center}
  \caption{\textbf{Overview of our two-stage method}. First, we learn temporally dense video representations in a self-supervised way, by exploiting the correspondence between the visual and auditory modalities of real videos. Second, the network is presented with real and fake data and is tasked with performing face forgery detection while simultaneously predicting, for the real videos, the representations learned in stage 1. We use many more real than fake samples, as the former are more easily acquired.}
\label{fig:teaser}
\end{figure}

Automatic face manipulation methods can realistically change someone's appearance or expression without requiring substantial human expertise or effort \cite{rossler2019faceforensics++, jiang2020deeperforensics, li2020advancing, li2020celeb, dolhansky2020deepfake}. This technology's potential social harm has spurred considerable research efforts to detect forgery content \cite{li2020face, haliassos2021lips, afchar2018mesonet, guera2018deepfake, zhou2017two, dang2020detection, mittal2020emotions, zhu2021face, zhao2021multi, chen2021magdr, fung2021deepfakeucl, qian2020thinking, gu2021spatiotemporal, khan2021video}. 

It is known that although deep learning-based detectors can achieve high accuracy on in-distribution data, performance often plummets on videos generated using novel manipulation methods (\textit{i.e.}, not seen during training) \cite{haliassos2021lips, li2020face, chai2020makes, li2020celeb, zhu2021face, wang2020cnn, cozzolino2018forensictransfer}. 

Various frame-based methods (\textit{i.e.}, that take a single frame as input) have been proposed to tackle cross-manipulation generalisation, including using data augmentation \cite{wang2020cnn}, truncating classifiers \cite{chai2020makes}, using 3D decomposition \cite{zhu2021face}, amplifying multi-band frequencies \cite{masi2020two}, and targeting the blending boundary between the background and the altered face \cite{li2020face}. Nevertheless, many still significantly underperform on novel forgery types or focus on low-level cues which can easily be corrupted by common perturbations like compression \cite{haliassos2021lips}.

It is reasonable to believe that incorporating the temporal dimension can improve performance, especially since many synthesis methods do not take into account temporal consistency during the generation process \cite{rossler2019faceforensics++}. However, as with frame-based methods, naively training deep networks on videos can lead to overfitting to the seen forgeries \cite{haliassos2021lips, sun2021improving, zheng2021exploring}. To counteract this, LipForensics \cite{haliassos2021lips} pre-trains on a large-scale lipreading dataset and then freezes part of the network to prevent it from focusing on low-level cues. It achieves strong performance in cross-manipulation generalisation and robustness to common corruptions. On the other hand, (1) it requires pre-training on a \textit{labelled} dataset, limiting its scalability; (2) it focuses exclusively on the mouth region; and (3) it freezes almost one third of the network when training on forgery data, which could sacrifice performance. A very recent method, FTCN \cite{zheng2021exploring}, demonstrates high cross-manipulation generalisation by constraining all spatial convolutional kernel sizes to one. But, as we show, the impressive generalisation may come at a cost of reduced robustness to compression changes.

\begin{figure}
\begin{center}
  \includegraphics[width=\linewidth]{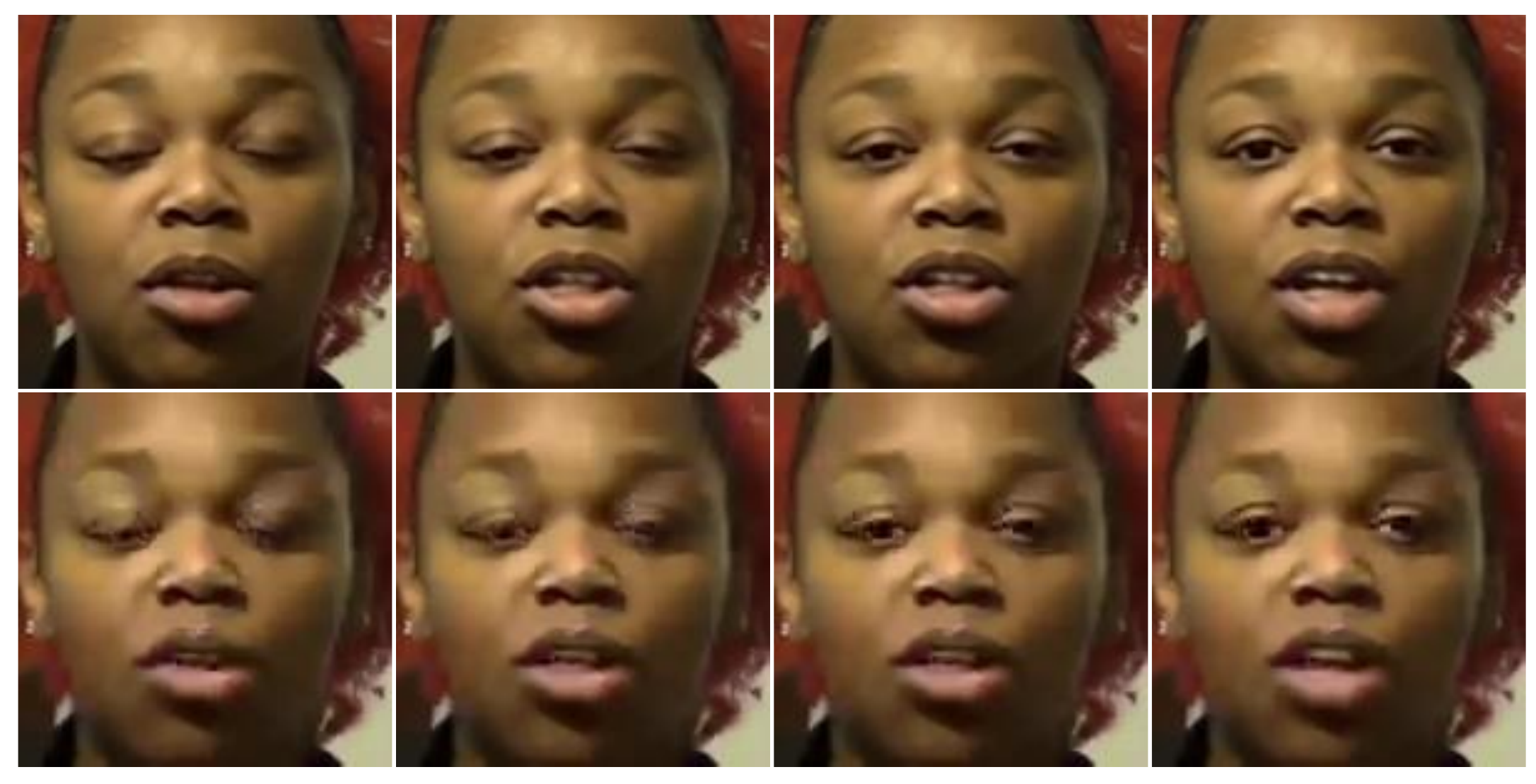}
\end{center}
  \caption{Top: consecutive frames of a fake video \cite{dolhansky2020deepfake}. Bottom: same frames but heavily compressed. High-level semantics remain largely undisturbed under compression.}
\label{fig:compression_imgs}
\end{figure}

In this work, we are motivated by the observation that fake videos often exhibit anomalous facial movements (including mouth, eyes, and brows) and expressions, as well as subtle changes in facial form over time. Such cues are high-level in nature and thus more resilient to corruptions which destroy low-level content, \textit{e.g.}, compression or blurring (see Figure \ref{fig:compression_imgs}). We ask ourselves whether it is possible to guide a detector to focus on such cues by utilising \textit{unlabelled} real videos, which are relatively easy to obtain with tools like face and voice activity detectors.

To this end, we propose a two-stage approach, termed \textit{RealForensics} (see Figure \ref{fig:teaser}). We first use self-supervision to exploit the known correspondence between the visual and auditory modalities in \textit{natural videos}. Inspired by the state-of-the-art method in image representation learning BYOL \cite{grill2020bootstrap}, we use a cross-modal student-teacher framework, where a student processing the video stream must predict representations formed by a slowly-improving teacher from the audio stream, and vice versa. We learn \textit{temporally dense} representations (one embedding per frame), since cues related to facial movements are often fast-varying. Our goal is to capture \textit{all} shared information between the two modalities, including factors associated with lexical content \cite{chung2016out}, emotion \cite{shukla2021does}, and identity \cite{nagrani2018learnable}. Hence, we \textit{directly predict} the teachers' outputs. In the second stage, the forgery detector is tasked with performing classification while simultaneously predicting video targets generated by the video student from the first stage. This prediction task incentivises the detector to focus on the aforementioned cues when classifying the samples and, as a result, alleviates overfitting.

Our contributions are as follows: (1) We present a novel two-stage detection approach that uses large amounts of natural talking faces for strong generalisation and robustness performance; this opens up the avenue for future forgery detection works to exploit the ubiquitous real videos online. (2) We propose, for the first stage, a non-contrastive self-supervised framework that learns temporally dense representations, and we validate its design for our task through ablations. (3) We achieve state-of-the-art performance in experiments that test cross-manipulation generalisation and robustness to common corruptions, and highlight the factors responsible for our method's performance.

\section{Related Works}

\subsection{Face forgery detection}
\begin{description}[wide,itemindent=\labelsep]
\item[General approaches.] Earlier works using convolutional neural networks (CNNs) include recasting steganalysis features as CNNs \cite{cozzolino2017recasting}, constraining convolutional filters \cite{bayar2016deep}, and using shallow networks \cite{afchar2018mesonet} to suppress high-level content. However, an unconstrained Xception \cite{chollet2017xception} network outperforms these approaches on more recent forgery types \cite{rossler2019faceforensics++}. Other works aim at detecting inconsistent head poses \cite{yang2019exposing} or irregular eye blinking \cite{li2018ictu}, although more recent fakes may not exhibit such anomalies. More recently, works have focused on attention mechanisms \cite{dang2020detection, zhao2021multi, wang2021representative}, exploiting the frequency spectrum \cite{durall2020watch, frank2020leveraging, qian2020thinking, luo2021generalizing, masi2020two, liu2021spatial, li2021frequency}, detecting anomalies in features from a face recognition network \cite{wang2019fakespotter}, or using extra identity information \cite{cozzolino2021id, agarwal2019protecting, dong2020identity}. \cite{fung2021deepfakeucl} and \cite{zhang2021deepfake} use self-supervision for frame-based detection, but do not study the effect of using many real samples.

\item[Cross-manipulation generalisation.] Detectors often generalise poorly to unseen forgeries \cite{cozzolino2018forensictransfer, li2020face, haliassos2021lips, chai2020makes, wang2020cnn}. Approaches to improve generalisation include applying augmentations \cite{wang2020cnn}, reconstructing the input as an auxiliary task \cite{cozzolino2018forensictransfer, du2019towards, nguyen2019multi}, mining frequency cues \cite{masi2020two, luo2021generalizing}, truncating classifiers \cite{chai2020makes}, focusing on self-consistency \cite{li2020face, zhao2021learning, li2018exposing, huh2018fighting}, or using spatio-temporal convolutional networks \cite{ganiyusufoglu2020spatio}. 

However, it has been shown that it is especially challenging to achieve cross-manipulation generalisation and at the same time perform well on corrupted data \cite{haliassos2021lips}. A closely related work to ours is LipForensics \cite{haliassos2021lips}, which addresses this by finetuning a network that was pre-trained to perform lipreading. Unlike our method, it requires a large-scale \textit{labelled} dataset and focuses exclusively on the mouth region. Recently, \cite{zheng2021exploring} report high generalisation by reducing the spatial kernel sizes of convolutional layers to 1, thus learning temporal inconsistencies while ignoring spatial ones. We target \textit{spatio-temporal} irregularities that may be more consistent with human perception of forgery cues. Finally, some works have focused on mismatches between the visual and auditory modalities in fake videos \cite{zhou2021joint, mittal2020emotions, agarwal2020detecting, korshunov2018speaker, chugh2020not}. Our work, on the other hand, is \textit{visual-only} at test-time: It uses the audio modality only for cross-modal supervision in an intermediate step, in which only real videos are used.

\end{description}

\subsection{Self-supervised learning}
\begin{description}[wide,itemindent=\labelsep]
\item[Image SSL.] Recently, contrastive learning using the InfoNCE loss \cite{oord2018representation} has become a popular approach in image representation learning \cite{chen2020simple, chen2020improved, he2020momentum, oord2018representation, wu2018unsupervised, henaff2020data, tian2020contrastive}. In this paradigm, the similarity between two views of an image is maximised, while different images (``negatives'') are repelled. Contrastive learning has also been used to learn dense visual representations \cite{wang2021dense, pinheiro2020unsupervised}. However, recent works that remove negatives generally outperform contrastive approaches \cite{caron2018deep, asano2019self, caron2020unsupervised, zbontar2021barlow, grill2020bootstrap, chen2021exploring, caron2021emerging}. Our work is partly inspired by BYOL \cite{grill2020bootstrap}, which uses a slowly-evolving teacher network that produces targets for a student to predict. The first stage of our approach can be viewed as an extension of BYOL to the audiovisual setting, in which we have a student-teacher pair for each modality and the output representations are temporally dense. Recent works \cite{niizumi2021byol} and \cite{recasens2021broaden, feichtenhofer2021large} also use BYOL-style training but are for audio-only learning and action recognition, respectively.

\item[Audiovisual SSL.] Many works exploit audiovisual correspondence for video action recognition \cite{arandjelovic2017look, arandjelovic2018objects, korbar2018cooperative, chung2016out, alwassel2019self, asano2020labelling, patrick2020multi, morgado2021audio, ma2020active}. However, these approaches learn a single representation for a video clip, which is less suitable for modelling the fine-grained movements of a speaking face. The very recent work \cite{ma2021contrastive} proposes to learn, in a contrastive manner, both global and local representations that are agnostic to the specific downstream task. In contrast, aside from methodological differences, our work focuses on cross-dataset generalisation and robustness for face forgery detection. Audiovisual methods have also been proposed for applications involving faces (\textit{e.g.}, audiovisual synchronisation and biometric matching). In general, methods that model lexical content tend to contrast samples from the same video for identity invariance \cite{chung2016out, chung2019perfect, chung2020seeing}. Conversely, works that learn identity embeddings tend to match misaligned video-audio pairs from the same person for invariance to lexical content \cite{nagrani2018seeing, nagrani2018learnable}. We posit that it is beneficial to capture both types of information for forgery detection and hence directly predict aligned embeddings.

\item[Generalisation via self-supervision.] It has been shown that using self-supervision as an auxiliary task, \textit{e.g.}, predicting rotations \cite{gidaris2018unsupervised} or solving jigsaw puzzles \cite{noroozi2016unsupervised}, can improve generalisation on the main task at hand \cite{hendrycks2019using, gidaris2019boosting, carlucci2019domain}. We use a similar idea for improving generalisation for forgery detection, but we \textit{learn} the targets in a separate stage before using them to define the auxiliary task.

\end{description}

\section{Method}

\begin{figure*}
\begin{center}
  \includegraphics[width=\linewidth]{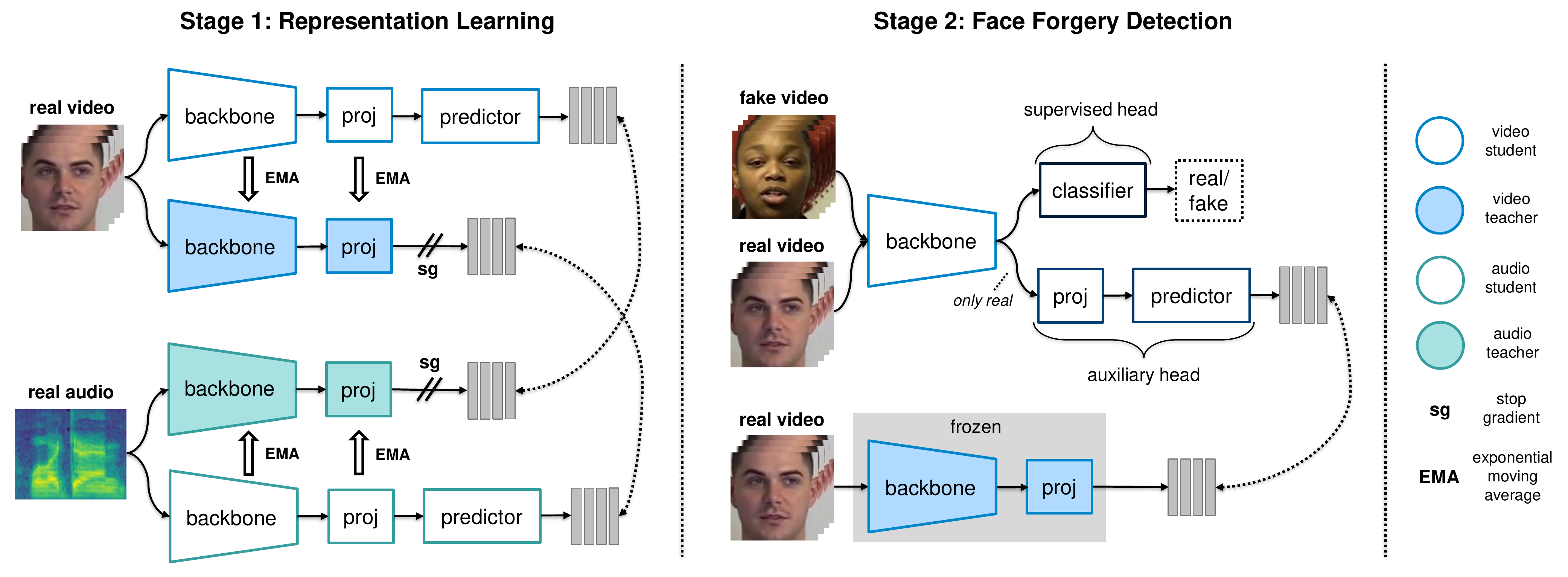}
\end{center}
  \caption{\textbf{The two stages of RealForensics}. In stage 1, the aim is to learn, in a self-supervised manner, frame-wise representations that capture information on natural facial behaviour and appearance. We utilise an audiovisual, cross-modal, student-teacher framework, whereby the student networks ingest real video and audio and try to predict the corresponding targets generated from the other modality. We also randomly mask the student inputs (omitted from the diagram for clarity). The teacher networks are momentum encoders that are updated via an exponential moving average (EMA), as in \cite{grill2020bootstrap}. In stage 2, the detector performs face forgery classification, while predicting the video targets produced by the (now frozen) video teacher from stage 1; only real videos contribute to the prediction loss. The video student from stage 1 is used to initialise the backbone. This multi-task formulation likely incentivises the network to detect forgeries based on stable cues that generalise well to unseen forgeries and are robust to low-level perturbations. Best viewed in colour.}
\label{fig:method}
\end{figure*}

RealForensics comprises two stages (see Figure \ref{fig:method}). The first stage involves learning temporally dense video representations using cross-modal self-supervision from many natural talking faces. These representations are subsequently used as prediction targets in the second stage to regularise the binary forgery classification task.

\subsection{Stage 1: representation learning}
Given real videos and the corresponding audio, we aim to learn video representations that capture information associated with facial appearance and behaviour. Cues like facial movements are temporally fine-grained by nature, and hence we wish to learn \textit{temporally dense} representations, \textit{i.e.}, an embedding per frame. We use a student-teacher framework without contrasting negatives for the following reasons. (1) This style of training has resulted in state-of-the-art results in image representation learning \cite{grill2020bootstrap}; (2) it incentivises the network to retain all information shared by the two modalities \cite{grill2020bootstrap}; and (3) it obviates the need for large batch sizes \cite{chen2020simple} or a queue \cite{he2020momentum} to store the negatives.

\begin{description}[wide,itemindent=\labelsep]

\item[Formulation.] We assume access to a large dataset $\mathcal{D}_r$ of real talking faces. A sample $x\in\mathcal{D}_r$ is a video $x^v\in\mathbb{R}^{T_v\times H\times W \times 3}$ (of $T_v$ video frames, height $H$, and width $W$) with its corresponding audio, represented as a log-mel spectrogram, $x^a\in\mathbb{R}^{T_a\times L}$ (of $T_a$ audio frames and $L$ mel filters). We ensure that $T_a=4T_v$.

Our architecture consists of a student and teacher pair for each modality. The teachers produce targets that the students from the other modality must predict. Specifically, teacher video and audio backbone networks, $f^v_t$ and $f^a_t$, produce embeddings $e^v_t=f^v_t(x^v)$ and $e^a_t=f^a_t(x^a)$ from the inputs, which are then passed through projectors, $g^v_t$ and $g^a_t$, to yield dense video and audio targets, $z^v_t=\text{norm}(g^v_t(e^v_t))\in\mathbb{R}^{T_v\times C}$ and $z^a_t=\text{norm}(g^a_t(e^a_t))\in\mathbb{R}^{T_v\times C}$, where $C$ is the dimensionality of the embeddings and $\text{norm}(\cdot)$ denotes $l_2$ normalisation across the channel dimension. Note that the audio backbone subsamples the temporal dimension such that the video and audio embeddings have the same shape. The students have the same architecture as their corresponding teachers, except that each student additionally contains a predictor, whose job is to predict the targets from the other modality. Let the video and audio predictions be $p^v=\text{norm}(h^v(z^v_s))$ and $p^a=\text{norm}(h^a(z^a_s))$, respectively, where $h^v$ and $h^a$ denote the predictors and $z^v_s$ and $z^a_s$ are the unnormalised student representations after the student projectors; then the loss is
\begin{align} \label{eq:prediction_loss}
    \mathcal{L}=\frac{1}{2}||\text{sg}\left(z^v_t\right)-p^a||^2_F+\frac{1}{2}||\text{sg}\left(z^a_t\right)-p^v||^2_F,
\end{align}
where $||\cdot||_F$ denotes the Frobenius norm, and sg, which stands for ``stop-gradient,'' emphasises that the targets are treated as constants. The total loss is averaged over all samples. The students are optimised via gradient descent, and the teachers are exponential moving averages of the students. That is, if we denote the video teacher weights as $\psi^v$ and the corresponding student weights as $\theta^v$, then at each iteration
\begin{align}
    \psi^v\leftarrow\mu\psi^v+(1-\mu)\theta^v,
\end{align}
where $\mu$ is a momentum parameter close to 1. The audio teacher weights are updated similarly.

\item[Transformer as predictor.] BYOL shows that the predictor is a necessary component to avoid \textit{representation collapse}, a situation where the representations for all samples are the same \cite{grill2020bootstrap}. We observe the same for our framework (see Section \ref{sec:ablations}). Whereas BYOL outputs global representations and thus uses an MLP as a predictor, we find that a shallow transformer is suitable for our dense representation learning task (see the appendix for an ablation).  

\item[Random masking.] We also find that random masking results in better representations (see Section \ref{sec:ablations}). For videos, we zero random rectangular regions in frames \cite{zhong2020random}, consistent across a whole video clip, as well as erase a random number of consecutive frames. For spectrograms, we erase a random number of consecutive audio frames and frequency bins. This is similar to the SpecAugment method \cite{park2019specaugment}, but without the time warping step. We apply this masking only to the inputs of the students. Intuitively, this forces the students to make use of context to infer the missing information and prevents them from overly relying on specific features of the input, \textit{e.g.}, the mouth region.

\item[Implementation details.] Unless specified otherwise, we use the following settings for this stage (see the appendix for more details).
\begin{itemize}[leftmargin=*]
    \item \textit{Inputs}. We extract the faces using face detection and alignment. A clip consists of 25 frames. The log-mel spectrograms contain 80 mel filters and 100 audio frames. During training, we randomly crop the video clips to size $140\times 140$ and resize them to $112\times 112$. We randomly apply horizontal flipping and grayscale transformation, each with probability 0.5. As mentioned, we also randomly mask the students' inputs.
    \item \textit{Backbones.} The video backbone is a Channel-Separated Convolutional Network (CSN) \cite{tran2019video}; we set the temporal strides to 1 to prevent temporal subsampling. The audio backbone is a ResNet18 \cite{he2016deep}, with the strides in the convolutional layers modified such that it subsamples the temporal dimension by 4, thus matching the temporal span of the video backbone's output.
    \item \textit{Projectors.} The projection network used for both the video and audio modalities is a single $1\times 1$ convolutional layer with output dimension of 256, followed by batch normalisation (BN) \cite{ioffe2015batch}.  We find that this BN layer helps with training, similarly to \cite{chen2021exploring}.
    \item \textit{Predictors.} The predictor for both modalities is a 1-block transformer encoder. It follows the design of a ViT block \cite{dosovitskiy2020image}. We use 8 attention heads, each with dimension 64, MLP dimension of 2048, and replace layer normalisation \cite{ba2016layer} with batch normalisation \cite{ioffe2015batch} before the MLP.
    \item \textit{Optimisation.} We use the AdamP optimiser \cite{heo2020adamp} with learning rate $7\times 10^{-4}$ and weight decay $10^{-2}$. We train for 150 epochs, with an initial 20-epoch linear warmup followed by a cosine decay schedule for the learning rate \cite{loshchilov2016sgdr}. The predictors' lr are kept fixed \cite{chen2021exploring}. The EMA momentum parameters for the teachers are set to 0.999.
\end{itemize}

\end{description}

\subsection{Stage 2: multi-task forgery detection}

The aim of this work is to learn a \textit{visual-only} forgery detector. Indeed, many forgery datasets do not officially release audio along with the videos \cite{rossler2019faceforensics++, jiang2020deeperforensics, li2020celeb}. As a result, at this stage we discard the audio student-teacher pair, after having served its purpose in stage 1. 

We propose to use the video teacher from stage 1 to produce targets for our network to predict. At the same time, the network performs forgery detection, in a multi-task fashion. Note that the teacher is frozen in this stage. Using this auxiliary loss likely encourages the network to classify real and fake videos by focusing on high-level spatio-temporal characteristics of facial appearance and behaviour. 

\begin{description}[wide,itemindent=\labelsep]
\item[Formulation.] We again use our dataset of real faces $\mathcal{D}_r$, but we now also assume access to a dataset of fake videos, $\mathcal{D}_f$.\footnote{In practice, the real samples for this stage include our auxiliary dataset as well as the real samples from the forgery dataset.} Our full dataset is thus $\mathcal{D}=\mathcal{D}_r \cup\mathcal{D}_f$. Our architecture consists of a shared backbone $f$ with weights $\theta_b$ and two heads: a supervised head with weights $\theta_s$ for the forgery classification loss and an auxiliary one $q$ with weights $\theta_a$ for the target prediction loss. The auxiliary loss is given by
\begin{align}
    \mathcal{L}_a(\mathcal{D}_r; \theta_b, \theta_a)=\mathop{\mathbb{E}}_{x\sim\mathcal{D}_r}||q\left(f(x^v;\theta_b);\theta_a\right)-t(x^v)||^2_F,
\end{align}
where $t$ is the teacher from stage 1, and the auxiliary head's and teacher's outputs are $l_2$-normalised as in stage 1.

The supervised loss $\mathcal{L}_s(\mathcal{D}; \theta_b, \theta_s)$ is a logit-adjusted version of binary cross entropy, as proposed in \cite{menon2020long}, to address any class imbalance (see the appendix for details). Moreoever, to obtain the logits, we $l_2$-normalise the feature vectors and the weights of the last linear layer (and set its bias to 0), thus obtaining a cosine classifier \cite{wang2017normface}. This combines better with the auxiliary loss, which can also be cast in terms of cosine similarity. Finally, the objective is given by
\begin{align}
    \min_{\theta_b,\theta_s, \theta_a}\mathcal{L}_s(\mathcal{D}; \theta_b, \theta_s) + w\mathcal{L}_a(\mathcal{D}_r; \theta_b, \theta_a),
\end{align}
where $w$ is a scaling factor, which we set to 1.

\item[Implementation details.] The video teacher is transferred from stage 1 and remains frozen henceforth. The backbone's architecture is the same as the video backbone in stage 1, and we initialise it with the learned weights. The auxiliary head is comprised of a randomly initialised projector and predictor as in stage 1. The supervised head is a cosine classifier, as previously described. A batch consists of 32 fake and 256 real samples, to effectively make use of the many more real samples available. We use the AdamP optimiser with learning rate $3\times 10^{-4}$ and the same preprocessing and augmentations described in stage 1. We train for 150 epochs and use the validation set for early stopping.

\end{description}

\section{Experiments}
\begin{description}[wide,itemindent=\labelsep]
\item[Auxiliary dataset.] We use the LRW dataset \cite{chung2016lip} without the labels for our extra real samples. It contains 500,000 videos of talking faces with hundreds of different identities. This dataset was also used by LipForensics \cite{haliassos2021lips}, which allows for fairer comparisons. In addition, its size strikes a balance between meaningful results and non-prohibitive computational costs. We present results for another dataset, VoxCeleb2 \cite{chung2018voxceleb2}, in Section \ref{sec:ablations}.
\item[Forgery datasets.] We use the following forgery datasets: (1) \textbf{FaceForensics++} (FF++) \cite{rossler2019faceforensics++} consists of 1,000 real videos and 4,000 fake videos, generated using two face swapping methods, Deepfakes \cite{deepfakes} and FaceSwap \cite{faceswap}, and two face reenactment methods, Face2Face \cite{thies2016face2face} and NeuralTextures \cite{thies2019deferred}. Unless stated otherwise, we use the mildly compressed version of the dataset (c23). As in \cite{haliassos2021lips, rossler2019faceforensics++}, we take the first 270 frames for each training video, and the first 110 frames for each validation/testing video. (2) \textbf{FaceShifter} \cite{li2020advancing} and (3) \textbf{DeeperForensics} \cite{jiang2020deeperforensics} are state-of-the-art face swapping methods that have been applied to the real videos of FF++; we use the test videos, according to the FF++ split. (4) \textbf{CelebDF-v2} \cite{li2020celeb} is a challenging face swapping dataset with 518 test videos. (5) \textbf{DFDC} is a subset of the Deepfake Detection Challenge Dataset (DFDC) \cite{dolhansky2020deepfake} used in \cite{haliassos2021lips}. It features 3,215 videos, many of which have been subjected to strong perturbations. 

\item[Evaluation metrics.] Following \textit{e.g.}, \cite{haliassos2021lips, zheng2021exploring, zhou2021joint, rossler2019faceforensics++, faceswap}, we use accuracy and area under the receiver operating characteristic curve (AUC) for evaluation. We use video-level metrics: For a single video we first uniformly sample non-overlapping clips and then average all clip predictions across the video.

\end{description}

\subsection{Cross-manipulation generalisation}
A deployed detector is expected to recognise fake videos that were created using methods \textit{not seen during training}, a non-trivial task in practice \cite{li2020face, haliassos2021lips, zheng2021exploring, nguyen2019multi}.  In this section, we follow the protocol used in \cite{haliassos2021lips, li2018exposing, nguyen2019multi} to evaluate our detector's ability to generalise to unseen manipulations.

Table \ref{table:ff_cross_manip} shows results obtained by RealForensics on each manipulation type in the FF++ dataset after training on the remaining types. Our detector works on par with the state-of-the-art without (1) using auxiliary labelled supervision \cite{haliassos2021lips}, (2) heavily constraining the network by freezing large parts \cite{haliassos2021lips} or removing spatial convolutions \cite{zheng2021exploring}, nor (3) using audio at test-time \cite{zhou2021joint}. We also outperform the baseline of training a CSN \cite{tran2019video} network on the forgery data (with the same augmentations as RealForensics), indicating the effectiveness of leveraging real data using our approach.

We also evaluate \textit{cross-dataset} generalisation by training on FF++ and then testing \textit{a single model} on unseen, challenging datasets: CelebDF-v2 \cite{li2020celeb}, DFDC \cite{dolhansky2020deepfake}, FaceShifter \cite{li2020advancing}, and DeeperForensics \cite{jiang2020deeperforensics}. The AUC results are given in Table \ref{table:cross_dataset}. Our method achieves state-of-the-art results on all datasets, suggesting that our detector performs well when exposed to more advanced forgeries than originally trained on. RealForensics also beats the CSN baseline by a large margin. Finally, as seen in Table \ref{table:params}, we achieve higher generalisation accuracy on FaceShifter and DeeperForensics than related methods, with fewer network parameters at test-time.

\begin{table}
\begin{center}
\begin{tabular}{l c c c c}\toprule
\multirow{2}{*}{Method} & \multicolumn{4}{c}{Train on remaining three} \\  
\cmidrule(lr){2-5}
& DF & FS & F2F & NT \\ \midrule
Xception \cite{rossler2019faceforensics++} & 93.9 & 51.2 & 86.8 & 79.7\\
CNN-aug \cite{wang2020cnn} & 87.5 & 56.3 & 80.1 & 67.8\\
Patch-based \cite{chai2020makes} & 94.0 & 60.5 & 87.3 & 84.8 \\
Face X-ray \cite{li2020face} & 99.5 & 93.2 & 94.5 & 92.5 \\
CNN-GRU \cite{sabir2019recurrent} & 97.6 & 47.6 & 85.8 & 86.6 \\
LipForensics \cite{haliassos2021lips} & 99.7 & 90.1 & \underline{99.7} & 99.1 \\
AV DFD \cite{zhou2021joint} & \underline{100.} & 90.5 & \underline{99.8} & 98.3 \\
FTCN \cite{zheng2021exploring} & 99.9 & \underline{99.9} & \underline{99.7} & \underline{99.2} \\ \midrule
CSN & 98.8 & 87.9 & 98.7 & 88.6 \\
RealForensics (ours) & \underline{100.} & \underline{97.1} & \underline{99.7} & \underline{99.2} \\ \bottomrule
\end{tabular}
\end{center}
\caption{\textbf{FF++ cross-manipulation generalisation.} AUC scores (\%) for each FF++ manipulation type after training on the remaining types. We use the test sets of Deepfakes (DF), FaceSwap (FS), Face2Face (F2F), and NeuralTextures (NT), as well as the real test videos. Top-2 best methods are \underline{underlined}.} 
\label{table:ff_cross_manip}
\end{table}

\begin{table}
\begin{center}
\resizebox{\linewidth}{!}{
\begin{tabular}{l c c c c | c}\toprule
Method & CDF & DFDC & FSh & DFo & Avg  \\ \midrule
Xception \cite{rossler2019faceforensics++} & 73.7 & 70.9 & 72.0 & 84.5 & 75.3  \\
CNN-aug \cite{wang2020cnn} & 75.6 & 72.1 & 65.7 & 74.4 & 72.0 \\
Patch-based \cite{chai2020makes} & 69.6 & 65.6 & 57.8 & 81.8 & 68.7 \\
Face X-ray \cite{li2020face} & 79.5 & 65.5 & 92.8 & 86.8 & 81.2 \\ 
CNN-GRU \cite{sabir2019recurrent} & 69.8 & 68.9 & 80.8 & 74.1 & 73.4 \\
Multi-task \cite{nguyen2019multi} & 75.7 & 68.1 & 66.0 & 77.7 & 71.9 \\
DSP-FWA \cite{li2019exposing} & 69.5 & 67.3 & 65.5 & 50.2 & 63.1 \\
Two-branch \cite{masi2020two} & 76.7 & --- & --- & --- & --- \\
LipForensics \cite{haliassos2021lips} & 82.4 & 73.5 & 97.1 & 97.6 & 87.7 \\
FTCN \cite{zheng2021exploring} & \textbf{86.9} & 74.0 & 98.8 & 98.8 & 89.6 \\ \midrule
CSN & 69.4 & 68.1 & 87.9 & 89.3 & 78.7 \\
RealForensics (ours) & \textbf{86.9} & \textbf{75.9} & \textbf{99.7} & \textbf{99.3} & \textbf{90.5} \\
\bottomrule 
\end{tabular}
}
\end{center}
\caption{\textbf{Cross-dataset generalisation.} AUC scores (\%) on CelebDF-v2 (CDF), DeepFake Detection Challenge (DFDC), FaceShifter (FSh), and DeeperForensics (DFo), after training on FaceForensics++. Best results are in \textbf{bold}.}
\label{table:cross_dataset}
\end{table}

\begin{table}
\begin{center}
\resizebox{\linewidth}{!}{
\begin{tabular}{l c c c c}\toprule
\multirow{2}{*}{Method} & \multicolumn{2}{c}{Settings} & \multicolumn{2}{c}{Accuracy} \\  
 \cmidrule(lr){2-3} \cmidrule(lr){4-5}
& Arch & \# params & FSh & DFo \\ \midrule
LipForensics \cite{haliassos2021lips} & RN+TCN \cite{martinez2020lipreading} & 36.0 & 87.5 & 90.4 \\
FTCN \cite{zheng2021exploring} & FTCN \cite{zheng2021exploring} & 26.6 & 93.9 & 91.1 \\
RealForensics (ours) & CSN \cite{tran2019video} & \textbf{21.4} & \textbf{97.1} & \textbf{97.1} \\ \bottomrule
\end{tabular}
}
\end{center}
\caption{\textbf{Parameters and generalisation accuracy.} Number of parameters (in millions), at test-time, for related state-of-the-art methods, and accuracy on FaceShifter (FSh) and DeeperForensics (DFo) after training on FaceForensics++. Best results are in \textbf{bold}.}
\label{table:params}
\end{table}

\subsection{Robustness to common corruptions}
\begin{figure}
\begin{center}
  \includegraphics[width=\linewidth]{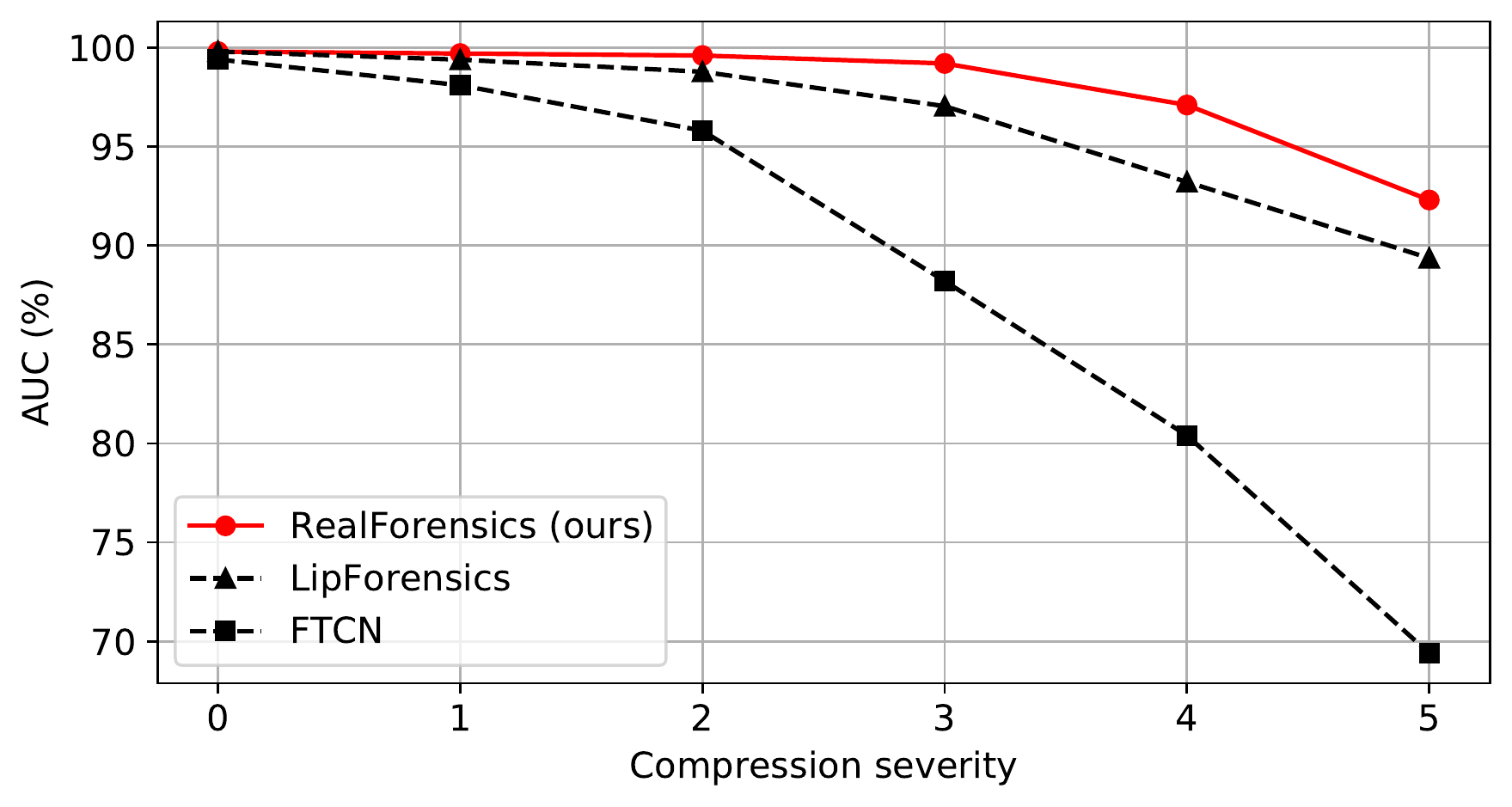}
\end{center}
  \caption{\textbf{Robustness to compression}. AUC scores (\%) on FaceForensics++ (FF++) at various H.264 video compression rates (23, 30, 32, 35, 38, 40), after training on FF++ with light compression (rate of 23).}
\label{fig:compression}
\end{figure}

In addition to good cross-manipulation generalisation, detectors should also be able to withstand common corruptions that videos may be subjected to on social media. We follow \cite{haliassos2021lips} to assess robustness to \textit{unseen} perturbations. As in \cite{haliassos2021lips}, we train on FF++ with grayscale clips and no augmentation other than horizontal flipping and random cropping, to avoid any intersection between train- and test-time perturbations. The set of perturbations, proposed in \cite{jiang2020deeperforensics}, are changes in saturation and contrast, block-wise occlusions, Gaussian noise and blur, pixelation and video compression. Each perturbation type is applied at five different intensity levels. Table \ref{table:robustness} presents the average AUC across all intensity levels for each corruption type. RealForensics suffers significantly less from common corruptions than frame-based methods that target low-level cues, such as \cite{li2020face, chai2020makes}, and also outperforms LipForensics and FTCN. (We use FTCN's publicly available model\footnote{\url{https://github.com/yinglinzheng/FTCN}}, which was trained on FF++ c23.) We notice that, relative to RealForensics and LipForensics, FTCN struggles on Gaussian noise and video compression (see also Figure \ref{fig:compression}), which disturb temporal coherence. This may be explained by FTCN's lack of spatial convolutions.

\begin{table*}
\begin{center}
\begin{tabular}{l | c | c c c c c c c | c}\toprule
Method & \textcolor{gray}{Clean} & Saturation & Contrast & Block & Noise & Blur & Pixel & Compress & Avg \\ \midrule
Xception \cite{rossler2019faceforensics++} & \textcolor{gray}{99.8} & 99.3 & 98.6 & \textbf{99.7} & 53.8 & 60.2 & 74.2 & 62.1 & 78.3 \\
CNN-aug \cite{wang2020cnn} & \textcolor{gray}{99.8} & 99.3 & 99.1 & 95.2 & 54.7 & 76.5 & 91.2 & 72.5 & 84.1 \\
Patch-based \cite{chai2020makes} & \textcolor{gray}{99.9} & 84.3 & 74.2 & 99.2 & 50.0 & 54.4 & 56.7 & 53.4 & 67.5  \\
Face X-ray \cite{li2020face} & \textcolor{gray}{99.8} & 97.6 & 88.5 & 99.1 & 49.8 & 63.8 & 88.6 & 55.2 & 77.5  \\
CNN-GRU \cite{sabir2019recurrent} & \textcolor{gray}{99.9} & 99.0 & 98.8 & 97.9 & 47.9 & 71.5 & 86.5 & 74.5 & 82.3 \\
LipForensics \cite{haliassos2021lips} & \textcolor{gray}{99.9} & \textbf{99.9} & \textbf{99.6} & 87.4 & 73.8 & \textbf{96.1} & 95.6 & 95.6 & 92.5 \\ 
FTCN \cite{zheng2021exploring} & \textcolor{gray}{99.4} & 99.4 & 96.7 & 97.1 & 53.1 & 95.8 & 98.2 & 86.4 & 89.5 \\ \midrule
RealForensics (ours) & \textcolor{gray}{99.8} & 99.8 & \textbf{99.6} & 98.9 & \textbf{79.7} & 95.3 & \textbf{98.4} & \textbf{97.6} & \textbf{95.6} \\
 \bottomrule
\end{tabular}
\end{center}
\caption{\textbf{Robustness to common corruptions.} Average AUC scores (\%) across five intensity levels for each corruption type proposed in \cite{jiang2020deeperforensics}. We also present, for each method, the average score across all corruptions. Best results are in \textbf{bold}. For a more detailed analysis, see the appendix.}
\label{table:robustness}
\end{table*}

\section{Ablations} \label{sec:ablations}
In this section, we present ablations to understand the factors responsible for our method's performance. See the appendix for more ablations.

\begin{description}[wide,itemindent=\labelsep]
\item[Framework ablation.] In Table \ref{table:framework_ablation}, we ablate different components of our method and inspect its generalisation performance on FaceShifter and DeeperForensics after training on FaceForensics++. We make the following observations. First, simply training a CSN \cite{tran2019video} model without our two-stage framework leads to a drop in accuracy of about $14\%$. Second, transferring the weights from stage 1 to the video backbone and finetuning the network on forgery data, without the auxiliary loss in stage 2, results in a drop of about $2\%$. This suggests that forcing the network to predict the video representations along with its main task has a positive regularisation effect. Finally, we observe modest improvements by employing logit adjustment \cite{menon2020long} for imbalanced classification and using time masking and random erasing \cite{zhong2020random}.

\item[Representation learning ablation.] For stage 1 of our method, we propose to learn temporally dense representations without contrasting negatives. Here, we test our choices against alternatives. We train the network with all combinations of the following settings: dense/global representations, with/without negatives, and with/without a predictor network. For global representation learning, we average-pool the output of the backbone networks, and use MLPs for the projector and predictor. To employ negatives, we use a queue of 65,536 samples and use the InfoNCE loss \cite{oord2018representation} with a temperature of 0.07. Note that global learning with negatives is similar to cross-modal contrastive learning used in \textit{e.g.}, \cite{morgado2021audio, ma2020active}. The predictor network used for global learning is an MLP and for dense learning a one-block transformer. Global learning with negatives and a predictor is analogous to the recent image representation learning method MoCo v3 \cite{chen2021empirical}, but for cross-modal learning. More information can be found in the appendix.

In Table \ref{table:representation_ablation}, we show accuracy scores on FaceShifter and DeeperForensics after training on FaceForensics++. We see that dense representations lead to significantly better performance than global. Also, consistent with the original BYOL method, we find that without negatives and without a predictor the outcome is \textit{representation collapse} \cite{grill2020bootstrap}. Without negatives and with global representations, no collapse was observed (with a predictor), but we had trouble achieving competitive performance. This may have to do with optimisation difficulties encountered without contrastive learning, since the subsequent inclusion of negatives yielded better results. However, adding negatives \textit{does not} seem to help when we use \textit{dense} learning (and a predictor).

\item[Effect of number of real samples.] Next, we vary the number of LRW samples in both stages of our method to see the effect on generalisation. As a baseline, we also consider simply treating the problem as an imbalanced classification task, \textit{i.e.}, training the model with logit adjustment (but without our proposed method). We can see in Figure \ref{fig:real_samples} that RealForensics benefits from a large number of real samples. Moreover, although generalisation for the baseline does increase with more real samples, the increase is significantly less than for RealForensics.

\begin{table}
\begin{center}
\begin{tabular}{l c c}\toprule
Method & FSh & DFo  \\ \midrule
RealForensics (ours) & \textbf{97.1} & \textbf{97.1} \\
\hspace{3mm} only CSN & 82.1 & 83.1 \\
\hspace{3mm} stage 1 + finetune & 95.0 & 95.2 \\
\hspace{3mm} w/o logit adjustment & 95.7 & 96.4 \\
\hspace{3mm} w/o time masking & 96.1 & 95.9 \\
\hspace{3mm} w/o random erasing & 96.3 & 96.3 \\
\bottomrule 
\end{tabular}
\end{center}
\caption{\textbf{Framework ablation.} Accuracy scores (\%) on FaceShifter (FSh) and DeeperForensics (DFo) after training on FaceForensics++. Refer to subsection ``Framework ablation'' for a discussion. Best results are in \textbf{bold}.}
\label{table:framework_ablation}
\end{table}

\begin{figure}
\begin{center}
  \includegraphics[width=\linewidth]{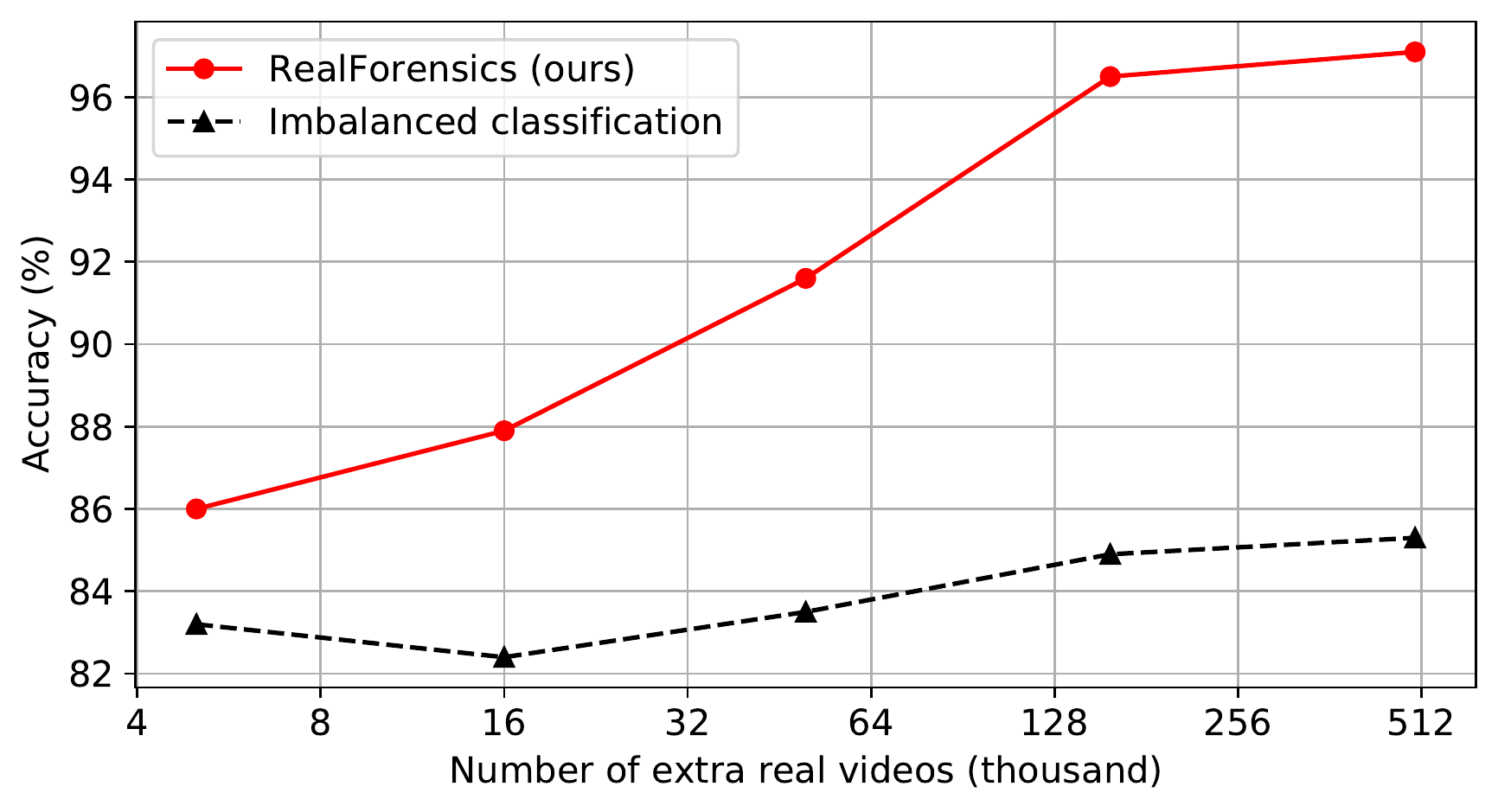}
\end{center}
  \caption{\textbf{\textbf{Effect of number of real samples}}. Accuracy scores (\%) as a function of the number of real samples from LRW, in log-scale. We show results for our method as well as a baseline where we treat the task as an imbalanced classification one. We average the accuracy of FaceShifter and DeeperForensics after training on FaceForensics++.}
\label{fig:real_samples}
\end{figure}

\begin{table}
\begin{center}
\begin{tabular}{c c c c c}\toprule
\multicolumn{3}{c}{Settings} & \multicolumn{2}{c}{Accuracy (\%)} \\ \cmidrule(lr){1-3} \cmidrule(lr){4-5}
Global/Dense & Negatives & Predictor & FSh & DFo  \\ \midrule
Global & \xmark & \xmark & n/a & n/a \\
Global & \xmark & \cmark & 70.7 & 74.1 \\
Global & \cmark & \xmark & 87.9 & 88.6 \\
Global & \cmark & \cmark & 87.9 & 89.1 \\
Dense & \xmark & \xmark & n/a & n/a \\
\rowcolor{light-gray}
Dense & \xmark & \cmark & \textbf{97.1} & \textbf{97.1} \\
Dense & \cmark & \xmark & 94.0 & 95.7 \\
Dense & \cmark & \cmark & 96.4 & 96.8 \\
\bottomrule 
\end{tabular}
\end{center}
\caption{\textbf{Representation learning ablation.} We ablate different components of our representation learning stage (stage 1). Note that ``n/a'' means that representation collapse was observed in stage 1. Refer to subsection ``Representation learning ablation'' for a discussion. Best results are in \textbf{bold}. Default setting is \colorbox{light-gray}{highlighted}.}
\label{table:representation_ablation}
\end{table}

\item[Using a different auxiliary dataset.] Here, we use the VoxCeleb2 dataset \cite{chung2018voxceleb2} for our extra real samples. It contains about 1 million videos of talking faces with various identities. We train with the same hyperparameters as for LRW. The AUC results (in \%) on CelebDF-v2, DFDC, FaceShifter, and DeeperForensics after training on FaceForensics++ are 82.9, 78.9, 99.3, and 98.8, respectively. This suggests that competitive results can be obtained with minimal tuning using a different dataset.

\end{description}

\section{Limitations / Societal Impact}
The strong generalisation of RealForensics comes at a cost of higher computational demands during training than methods that do not use auxiliary datasets; however, this is not the case at test-time. Moreover, our detector takes videos as input, and thus does not work for single images. Despite state-of-the-art accuracy, we also observe that when our network produces wrong predictions, they are often confidently wrong, so the probabilities outputted by the model should be interpreted with care. This issue of \textit{model calibration} is common in deep learning models \cite{guo2017calibration}, including forgery detectors; thus, an important future direction would be to apply methods in calibration literature \cite{guo2017calibration} to detectors. 

Although the purpose of research on forgery detection is to protect society, there are a few concerns that should be kept in mind. For example, pointing out the flaws in current face forgeries could facilitate the development of even better fake videos in the future. This, however, is less of an issue for methods that do not target a \textit{specific} cue, such as RealForensics. Further, it is not prudent for a deployed system to rely exclusively on a single detection method. For greater effectiveness, it should employ an ensemble of independent approaches.

\section{Conclusion}
In this paper, we propose RealForensics, an approach that uses large amounts of unlabelled real data to detect fake videos. We have shown that our method simultaneously achieves strong cross-manipulation generalisation and robustness to common corruptions. We hope our study encourages future research on leveraging real faces for robust forgery detection.

\begin{description}[wide,itemindent=\labelsep]
\item[Acknowledgements.] We thank Konstantinos Vougioukas for fruitful discussions. This work has been supported in part by Meta Platforms through research funding made available directly to Imperial College London (project P93445: Cross-modal learning of emotions). Alexandros Haliassos was financially supported by an Imperial President's PhD Scholarship. All training, testing, and ablation studies have been conducted at Imperial College. 
\end{description}

\newpage

{\small
\bibliographystyle{ieee_fullname}
\bibliography{egbib}

\begin{thebibliography}{100}\itemsep=-1pt

\bibitem{deepfakes}
Deepfakes.
\newblock \url{https://github.com/deepfakes/faceswap}.
\newblock [Accessed: 2020-11-12].

\bibitem{faceswap}
Faceswap.
\newblock \url{https://github.com/MarekKowalski/FaceSwap}.
\newblock [Accessed: 2020-11-12].

\bibitem{afchar2018mesonet}
Darius Afchar, Vincent Nozick, Junichi Yamagishi, and Isao Echizen.
\newblock Mesonet: a compact facial video forgery detection network.
\newblock In {\em 2018 IEEE International Workshop on Information Forensics and
  Security (WIFS)}, pages 1--7. IEEE, 2018.

\bibitem{afouras2020self}
Triantafyllos Afouras, Andrew Owens, Joon~Son Chung, and Andrew Zisserman.
\newblock Self-supervised learning of audio-visual objects from video.
\newblock In {\em Computer Vision--ECCV 2020: 16th European Conference,
  Glasgow, UK, August 23--28, 2020, Proceedings, Part XVIII 16}, pages
  208--224. Springer, 2020.

\bibitem{agarwal2020detecting}
Shruti Agarwal, Hany Farid, Ohad Fried, and Maneesh Agrawala.
\newblock Detecting deep-fake videos from phoneme-viseme mismatches.
\newblock In {\em Proceedings of the IEEE/CVF Conference on Computer Vision and
  Pattern Recognition Workshops}, pages 660--661, 2020.

\bibitem{agarwal2019protecting}
Shruti Agarwal, Hany Farid, Yuming Gu, Mingming He, Koki Nagano, and Hao Li.
\newblock Protecting world leaders against deep fakes.
\newblock In {\em CVPR workshops}, volume~1, 2019.

\bibitem{alwassel2019self}
Humam Alwassel, Dhruv Mahajan, Bruno Korbar, Lorenzo Torresani, Bernard Ghanem,
  and Du Tran.
\newblock Self-supervised learning by cross-modal audio-video clustering.
\newblock {\em arXiv preprint arXiv:1911.12667}, 2019.

\bibitem{arandjelovic2017look}
Relja Arandjelovic and Andrew Zisserman.
\newblock Look, listen and learn.
\newblock In {\em Proceedings of the IEEE International Conference on Computer
  Vision}, pages 609--617, 2017.

\bibitem{arandjelovic2018objects}
Relja Arandjelovic and Andrew Zisserman.
\newblock Objects that sound.
\newblock In {\em Proceedings of the European conference on computer vision
  (ECCV)}, pages 435--451, 2018.

\bibitem{asano2020labelling}
Yuki~M Asano, Mandela Patrick, Christian Rupprecht, and Andrea Vedaldi.
\newblock Labelling unlabelled videos from scratch with multi-modal
  self-supervision.
\newblock {\em arXiv preprint arXiv:2006.13662}, 2020.

\bibitem{asano2019self}
Yuki~Markus Asano, Christian Rupprecht, and Andrea Vedaldi.
\newblock Self-labelling via simultaneous clustering and representation
  learning.
\newblock {\em arXiv preprint arXiv:1911.05371}, 2019.

\bibitem{ba2016layer}
Jimmy~Lei Ba, Jamie~Ryan Kiros, and Geoffrey~E Hinton.
\newblock Layer normalization.
\newblock {\em arXiv preprint arXiv:1607.06450}, 2016.

\bibitem{bayar2016deep}
Belhassen Bayar and Matthew~C Stamm.
\newblock A deep learning approach to universal image manipulation detection
  using a new convolutional layer.
\newblock In {\em Proceedings of the 4th ACM workshop on information hiding and
  multimedia security}, pages 5--10, 2016.

\bibitem{bulat2017far}
Adrian Bulat and Georgios Tzimiropoulos.
\newblock How far are we from solving the 2d \& 3d face alignment problem? (and
  a dataset of 230,000 3d facial landmarks).
\newblock In {\em International Conference on Computer Vision}, 2017.

\bibitem{carlucci2019domain}
Fabio~M Carlucci, Antonio D'Innocente, Silvia Bucci, Barbara Caputo, and
  Tatiana Tommasi.
\newblock Domain generalization by solving jigsaw puzzles.
\newblock In {\em Proceedings of the IEEE/CVF Conference on Computer Vision and
  Pattern Recognition}, pages 2229--2238, 2019.

\bibitem{caron2018deep}
Mathilde Caron, Piotr Bojanowski, Armand Joulin, and Matthijs Douze.
\newblock Deep clustering for unsupervised learning of visual features.
\newblock In {\em Proceedings of the European Conference on Computer Vision
  (ECCV)}, pages 132--149, 2018.

\bibitem{caron2020unsupervised}
Mathilde Caron, Ishan Misra, Julien Mairal, Priya Goyal, Piotr Bojanowski, and
  Armand Joulin.
\newblock Unsupervised learning of visual features by contrasting cluster
  assignments.
\newblock {\em arXiv preprint arXiv:2006.09882}, 2020.

\bibitem{caron2021emerging}
Mathilde Caron, Hugo Touvron, Ishan Misra, Herv{\'e} J{\'e}gou, Julien Mairal,
  Piotr Bojanowski, and Armand Joulin.
\newblock Emerging properties in self-supervised vision transformers.
\newblock {\em arXiv preprint arXiv:2104.14294}, 2021.

\bibitem{chai2020makes}
Lucy Chai, David Bau, Ser-Nam Lim, and Phillip Isola.
\newblock What makes fake images detectable? understanding properties that
  generalize.
\newblock In {\em European Conference on Computer Vision}, pages 103--120.
  Springer, 2020.

\bibitem{chen2020simple}
Ting Chen, Simon Kornblith, Mohammad Norouzi, and Geoffrey Hinton.
\newblock A simple framework for contrastive learning of visual
  representations.
\newblock In {\em International conference on machine learning}, pages
  1597--1607. PMLR, 2020.

\bibitem{chen2020improved}
Xinlei Chen, Haoqi Fan, Ross Girshick, and Kaiming He.
\newblock Improved baselines with momentum contrastive learning.
\newblock {\em arXiv preprint arXiv:2003.04297}, 2020.

\bibitem{chen2021exploring}
Xinlei Chen and Kaiming He.
\newblock Exploring simple siamese representation learning.
\newblock In {\em Proceedings of the IEEE/CVF Conference on Computer Vision and
  Pattern Recognition}, pages 15750--15758, 2021.

\bibitem{chen2021empirical}
Xinlei Chen, Saining Xie, and Kaiming He.
\newblock An empirical study of training self-supervised vision transformers.
\newblock {\em arXiv preprint arXiv:2104.02057}, 2021.

\bibitem{chen2021magdr}
Zhikai Chen, Lingxi Xie, Shanmin Pang, Yong He, and Bo Zhang.
\newblock Magdr: Mask-guided detection and reconstruction for defending
  deepfakes.
\newblock In {\em Proceedings of the IEEE/CVF Conference on Computer Vision and
  Pattern Recognition}, pages 9014--9023, 2021.

\bibitem{chollet2017xception}
Fran{\c{c}}ois Chollet.
\newblock Xception: Deep learning with depthwise separable convolutions.
\newblock In {\em Proceedings of the IEEE conference on computer vision and
  pattern recognition}, pages 1251--1258, 2017.

\bibitem{chugh2020not}
Komal Chugh, Parul Gupta, Abhinav Dhall, and Ramanathan Subramanian.
\newblock Not made for each other-audio-visual dissonance-based deepfake
  detection and localization.
\newblock In {\em Proceedings of the 28th ACM International Conference on
  Multimedia}, pages 439--447, 2020.

\bibitem{chung2018voxceleb2}
Joon~Son Chung, Arsha Nagrani, and Andrew Zisserman.
\newblock Voxceleb2: Deep speaker recognition.
\newblock {\em arXiv preprint arXiv:1806.05622}, 2018.

\bibitem{chung2016lip}
Joon~Son Chung and Andrew Zisserman.
\newblock Lip reading in the wild.
\newblock In {\em Asian Conference on Computer Vision}, pages 87--103.
  Springer, 2016.

\bibitem{chung2016out}
Joon~Son Chung and Andrew Zisserman.
\newblock Out of time: automated lip sync in the wild.
\newblock In {\em Asian conference on computer vision}, pages 251--263.
  Springer, 2016.

\bibitem{chung2019perfect}
Soo-Whan Chung, Joon~Son Chung, and Hong-Goo Kang.
\newblock Perfect match: Improved cross-modal embeddings for audio-visual
  synchronisation.
\newblock In {\em ICASSP 2019-2019 IEEE International Conference on Acoustics,
  Speech and Signal Processing (ICASSP)}, pages 3965--3969. IEEE, 2019.

\bibitem{chung2020seeing}
Soo-Whan Chung, Hong~Goo Kang, and Joon~Son Chung.
\newblock Seeing voices and hearing voices: learning discriminative embeddings
  using cross-modal self-supervision.
\newblock {\em arXiv preprint arXiv:2004.14326}, 2020.

\bibitem{cozzolino2017recasting}
Davide Cozzolino, Giovanni Poggi, and Luisa Verdoliva.
\newblock Recasting residual-based local descriptors as convolutional neural
  networks: an application to image forgery detection.
\newblock In {\em Proceedings of the 5th ACM Workshop on Information Hiding and
  Multimedia Security}, pages 159--164, 2017.

\bibitem{cozzolino2021id}
Davide Cozzolino, Andreas Rossler, Justus Thies, Matthias Nie{\ss}ner, and
  Luisa Verdoliva.
\newblock Id-reveal: Identity-aware deepfake video detection.
\newblock In {\em Proceedings of the IEEE/CVF International Conference on
  Computer Vision}, pages 15108--15117, 2021.

\bibitem{cozzolino2018forensictransfer}
Davide Cozzolino, Justus Thies, Andreas R{\"o}ssler, Christian Riess, Matthias
  Nie{\ss}ner, and Luisa Verdoliva.
\newblock Forensictransfer: Weakly-supervised domain adaptation for forgery
  detection.
\newblock {\em arXiv preprint arXiv:1812.02510}, 2018.

\bibitem{dang2020detection}
Hao Dang, Feng Liu, Joel Stehouwer, Xiaoming Liu, and Anil~K Jain.
\newblock On the detection of digital face manipulation.
\newblock In {\em Proceedings of the IEEE/CVF Conference on Computer Vision and
  Pattern recognition}, pages 5781--5790, 2020.

\bibitem{deng2020retinaface}
Jiankang Deng, Jia Guo, Evangelos Ververas, Irene Kotsia, and Stefanos
  Zafeiriou.
\newblock Retinaface: Single-shot multi-level face localisation in the wild.
\newblock In {\em Proceedings of the IEEE/CVF Conference on Computer Vision and
  Pattern Recognition}, pages 5203--5212, 2020.

\bibitem{dolhansky2020deepfake}
Brian Dolhansky, Joanna Bitton, Ben Pflaum, Jikuo Lu, Russ Howes, Menglin Wang,
  and Cristian~Canton Ferrer.
\newblock The deepfake detection challenge (dfdc) dataset.
\newblock {\em arXiv preprint arXiv:2006.07397}, 2020.

\bibitem{dong2020identity}
Xiaoyi Dong, Jianmin Bao, Dongdong Chen, Weiming Zhang, Nenghai Yu, Dong Chen,
  Fang Wen, and Baining Guo.
\newblock Identity-driven deepfake detection.
\newblock {\em arXiv preprint arXiv:2012.03930}, 2020.

\bibitem{dosovitskiy2020image}
Alexey Dosovitskiy, Lucas Beyer, Alexander Kolesnikov, Dirk Weissenborn,
  Xiaohua Zhai, Thomas Unterthiner, Mostafa Dehghani, Matthias Minderer, Georg
  Heigold, Sylvain Gelly, et~al.
\newblock An image is worth 16x16 words: Transformers for image recognition at
  scale.
\newblock {\em arXiv preprint arXiv:2010.11929}, 2020.

\bibitem{du2019towards}
Mengnan Du, Shiva Pentyala, Yuening Li, and Xia Hu.
\newblock Towards generalizable forgery detection with locality-aware
  autoencoder.
\newblock {\em arXiv e-prints}, pages arXiv--1909, 2019.

\bibitem{durall2020watch}
Ricard Durall, Margret Keuper, and Janis Keuper.
\newblock Watch your up-convolution: Cnn based generative deep neural networks
  are failing to reproduce spectral distributions.
\newblock In {\em Proceedings of the IEEE/CVF Conference on Computer Vision and
  Pattern Recognition}, pages 7890--7899, 2020.

\bibitem{feichtenhofer2021large}
Christoph Feichtenhofer, Haoqi Fan, Bo Xiong, Ross Girshick, and Kaiming He.
\newblock A large-scale study on unsupervised spatiotemporal representation
  learning.
\newblock In {\em Proceedings of the IEEE/CVF Conference on Computer Vision and
  Pattern Recognition}, pages 3299--3309, 2021.

\bibitem{frank2020leveraging}
Joel Frank, Thorsten Eisenhofer, Lea Sch{\"o}nherr, Asja Fischer, Dorothea
  Kolossa, and Thorsten Holz.
\newblock Leveraging frequency analysis for deep fake image recognition.
\newblock In {\em International Conference on Machine Learning}, pages
  3247--3258. PMLR, 2020.

\bibitem{fung2021deepfakeucl}
Sheldon Fung, Xuequan Lu, Chao Zhang, and Chang-Tsun Li.
\newblock Deepfakeucl: Deepfake detection via unsupervised contrastive
  learning.
\newblock {\em arXiv preprint arXiv:2104.11507}, 2021.

\bibitem{ganiyusufoglu2020spatio}
Ipek Ganiyusufoglu, L~Minh Ng{\^o}, Nedko Savov, Sezer Karaoglu, and Theo
  Gevers.
\newblock Spatio-temporal features for generalized detection of deepfake
  videos.
\newblock {\em arXiv preprint arXiv:2010.11844}, 2020.

\bibitem{gidaris2019boosting}
Spyros Gidaris, Andrei Bursuc, Nikos Komodakis, Patrick P{\'e}rez, and Matthieu
  Cord.
\newblock Boosting few-shot visual learning with self-supervision.
\newblock In {\em Proceedings of the IEEE/CVF International Conference on
  Computer Vision}, pages 8059--8068, 2019.

\bibitem{gidaris2018unsupervised}
Spyros Gidaris, Praveer Singh, and Nikos Komodakis.
\newblock Unsupervised representation learning by predicting image rotations.
\newblock {\em arXiv preprint arXiv:1803.07728}, 2018.

\bibitem{grill2020bootstrap}
Jean-Bastien Grill, Florian Strub, Florent Altch{\'e}, Corentin Tallec,
  Pierre~H Richemond, Elena Buchatskaya, Carl Doersch, Bernardo~Avila Pires,
  Zhaohan~Daniel Guo, Mohammad~Gheshlaghi Azar, et~al.
\newblock Bootstrap your own latent: A new approach to self-supervised
  learning.
\newblock {\em arXiv preprint arXiv:2006.07733}, 2020.

\bibitem{gu2021spatiotemporal}
Zhihao Gu, Yang Chen, Taiping Yao, Shouhong Ding, Jilin Li, Feiyue Huang, and
  Lizhuang Ma.
\newblock Spatiotemporal inconsistency learning for deepfake video detection.
\newblock In {\em Proceedings of the 29th ACM International Conference on
  Multimedia}, pages 3473--3481, 2021.

\bibitem{guera2018deepfake}
David G{\"u}era and Edward~J Delp.
\newblock Deepfake video detection using recurrent neural networks.
\newblock In {\em 2018 15th IEEE international conference on advanced video and
  signal based surveillance (AVSS)}, pages 1--6. IEEE, 2018.

\bibitem{guo2017calibration}
Chuan Guo, Geoff Pleiss, Yu Sun, and Kilian~Q Weinberger.
\newblock On calibration of modern neural networks.
\newblock In {\em International Conference on Machine Learning}, pages
  1321--1330. PMLR, 2017.

\bibitem{haliassossupplementary}
Alexandros Haliassos, Konstantinos Vougioukas, Stavros Petridis, and Maja
  Pantic.
\newblock Supplementary material for lips don’t lie: A generalisable and
  robust approach to face forgery detection.

\bibitem{haliassos2021lips}
Alexandros Haliassos, Konstantinos Vougioukas, Stavros Petridis, and Maja
  Pantic.
\newblock Lips don't lie: A generalisable and robust approach to face forgery
  detection.
\newblock In {\em Proceedings of the IEEE/CVF Conference on Computer Vision and
  Pattern Recognition}, pages 5039--5049, 2021.

\bibitem{he2020momentum}
Kaiming He, Haoqi Fan, Yuxin Wu, Saining Xie, and Ross Girshick.
\newblock Momentum contrast for unsupervised visual representation learning.
\newblock In {\em Proceedings of the IEEE/CVF Conference on Computer Vision and
  Pattern Recognition}, pages 9729--9738, 2020.

\bibitem{he2016deep}
Kaiming He, Xiangyu Zhang, Shaoqing Ren, and Jian Sun.
\newblock Deep residual learning for image recognition.
\newblock In {\em Proceedings of the IEEE conference on computer vision and
  pattern recognition}, pages 770--778, 2016.

\bibitem{he2021forgerynet}
Yinan He, Bei Gan, Siyu Chen, Yichun Zhou, Guojun Yin, Luchuan Song, Lu Sheng,
  Jing Shao, and Ziwei Liu.
\newblock Forgerynet: A versatile benchmark for comprehensive forgery analysis.
\newblock In {\em Proceedings of the IEEE/CVF Conference on Computer Vision and
  Pattern Recognition}, pages 4360--4369, 2021.

\bibitem{henaff2020data}
Olivier Henaff.
\newblock Data-efficient image recognition with contrastive predictive coding.
\newblock In {\em International Conference on Machine Learning}, pages
  4182--4192. PMLR, 2020.

\bibitem{hendrycks2019using}
Dan Hendrycks, Mantas Mazeika, Saurav Kadavath, and Dawn Song.
\newblock Using self-supervised learning can improve model robustness and
  uncertainty.
\newblock {\em arXiv preprint arXiv:1906.12340}, 2019.

\bibitem{heo2020adamp}
Byeongho Heo, Sanghyuk Chun, Seong~Joon Oh, Dongyoon Han, Sangdoo Yun, Gyuwan
  Kim, Youngjung Uh, and Jung-Woo Ha.
\newblock Adamp: Slowing down the slowdown for momentum optimizers on
  scale-invariant weights.
\newblock {\em arXiv preprint arXiv:2006.08217}, 2020.

\bibitem{huh2018fighting}
Minyoung Huh, Andrew Liu, Andrew Owens, and Alexei~A Efros.
\newblock Fighting fake news: Image splice detection via learned
  self-consistency.
\newblock In {\em Proceedings of the European Conference on Computer Vision
  (ECCV)}, pages 101--117, 2018.

\bibitem{ioffe2015batch}
Sergey Ioffe and Christian Szegedy.
\newblock Batch normalization: Accelerating deep network training by reducing
  internal covariate shift.
\newblock In {\em International conference on machine learning}, pages
  448--456. PMLR, 2015.

\bibitem{jiang2020deeperforensics}
Liming Jiang, Ren Li, Wayne Wu, Chen Qian, and Chen~Change Loy.
\newblock Deeperforensics-1.0: A large-scale dataset for real-world face
  forgery detection.
\newblock In {\em Proceedings of the IEEE/CVF Conference on Computer Vision and
  Pattern Recognition}, pages 2889--2898, 2020.

\bibitem{khan2021video}
Sohail~Ahmed Khan and Hang Dai.
\newblock Video transformer for deepfake detection with incremental learning.
\newblock In {\em Proceedings of the 29th ACM International Conference on
  Multimedia}, pages 1821--1828, 2021.

\bibitem{korbar2018cooperative}
Bruno Korbar, Du Tran, and Lorenzo Torresani.
\newblock Cooperative learning of audio and video models from self-supervised
  synchronization.
\newblock {\em arXiv preprint arXiv:1807.00230}, 2018.

\bibitem{korshunov2018speaker}
Pavel Korshunov and S{\'e}bastien Marcel.
\newblock Speaker inconsistency detection in tampered video.
\newblock In {\em 2018 26th European signal processing conference (EUSIPCO)},
  pages 2375--2379. IEEE, 2018.

\bibitem{li2021frequency}
Jiaming Li, Hongtao Xie, Jiahong Li, Zhongyuan Wang, and Yongdong Zhang.
\newblock Frequency-aware discriminative feature learning supervised by
  single-center loss for face forgery detection.
\newblock In {\em Proceedings of the IEEE/CVF Conference on Computer Vision and
  Pattern Recognition}, pages 6458--6467, 2021.

\bibitem{li2020advancing}
Lingzhi Li, Jianmin Bao, Hao Yang, Dong Chen, and Fang Wen.
\newblock Advancing high fidelity identity swapping for forgery detection.
\newblock In {\em Proceedings of the IEEE/CVF Conference on Computer Vision and
  Pattern Recognition}, pages 5074--5083, 2020.

\bibitem{li2020face}
Lingzhi Li, Jianmin Bao, Ting Zhang, Hao Yang, Dong Chen, Fang Wen, and Baining
  Guo.
\newblock Face x-ray for more general face forgery detection.
\newblock In {\em Proceedings of the IEEE/CVF Conference on Computer Vision and
  Pattern Recognition}, pages 5001--5010, 2020.

\bibitem{li2018ictu}
Yuezun Li, Ming-Ching Chang, and Siwei Lyu.
\newblock In ictu oculi: Exposing ai created fake videos by detecting eye
  blinking.
\newblock In {\em 2018 IEEE International Workshop on Information Forensics and
  Security (WIFS)}, pages 1--7. IEEE, 2018.

\bibitem{li2018exposing}
Yuezun Li and Siwei Lyu.
\newblock Exposing deepfake videos by detecting face warping artifacts.
\newblock {\em arXiv preprint arXiv:1811.00656}, 2018.

\bibitem{li2019exposing}
Yuezun Li and Siwei Lyu.
\newblock Exposing deepfake videos by detecting face warping artifacts.
\newblock In {\em IEEE Conference on Computer Vision and Pattern Recognition
  Workshops (CVPRW)}, 2019.

\bibitem{li2020celeb}
Yuezun Li, Xin Yang, Pu Sun, Honggang Qi, and Siwei Lyu.
\newblock Celeb-df: A large-scale challenging dataset for deepfake forensics.
\newblock In {\em Proceedings of the IEEE/CVF Conference on Computer Vision and
  Pattern Recognition}, pages 3207--3216, 2020.

\bibitem{liu2021spatial}
Honggu Liu, Xiaodan Li, Wenbo Zhou, Yuefeng Chen, Yuan He, Hui Xue, Weiming
  Zhang, and Nenghai Yu.
\newblock Spatial-phase shallow learning: rethinking face forgery detection in
  frequency domain.
\newblock In {\em Proceedings of the IEEE/CVF Conference on Computer Vision and
  Pattern Recognition}, pages 772--781, 2021.

\bibitem{loshchilov2016sgdr}
Ilya Loshchilov and Frank Hutter.
\newblock Sgdr: Stochastic gradient descent with warm restarts.
\newblock {\em arXiv preprint arXiv:1608.03983}, 2016.

\bibitem{luo2021generalizing}
Yuchen Luo, Yong Zhang, Junchi Yan, and Wei Liu.
\newblock Generalizing face forgery detection with high-frequency features.
\newblock In {\em Proceedings of the IEEE/CVF Conference on Computer Vision and
  Pattern Recognition}, pages 16317--16326, 2021.

\bibitem{ma2020active}
Shuang Ma, Zhaoyang Zeng, Daniel McDuff, and Yale Song.
\newblock Active contrastive learning of audio-visual video representations.
\newblock {\em arXiv preprint arXiv:2009.09805}, 2020.

\bibitem{ma2021contrastive}
Shuang Ma, Zhaoyang Zeng, Daniel McDuff, and Yale Song.
\newblock Contrastive learning of global and local audio-visual
  representations.
\newblock {\em arXiv preprint arXiv:2104.05418}, 2021.

\bibitem{martinez2020lipreading}
Brais Martinez, Pingchuan Ma, Stavros Petridis, and Maja Pantic.
\newblock Lipreading using temporal convolutional networks.
\newblock In {\em ICASSP 2020-2020 IEEE International Conference on Acoustics,
  Speech and Signal Processing (ICASSP)}, pages 6319--6323. IEEE, 2020.

\bibitem{masi2020two}
Iacopo Masi, Aditya Killekar, Royston~Marian Mascarenhas, Shenoy~Pratik
  Gurudatt, and Wael AbdAlmageed.
\newblock Two-branch recurrent network for isolating deepfakes in videos.
\newblock In {\em European Conference on Computer Vision}, pages 667--684.
  Springer, 2020.

\bibitem{menon2020long}
Aditya~Krishna Menon, Sadeep Jayasumana, Ankit~Singh Rawat, Himanshu Jain,
  Andreas Veit, and Sanjiv Kumar.
\newblock Long-tail learning via logit adjustment.
\newblock {\em arXiv preprint arXiv:2007.07314}, 2020.

\bibitem{mittal2020emotions}
Trisha Mittal, Uttaran Bhattacharya, Rohan Chandra, Aniket Bera, and Dinesh
  Manocha.
\newblock Emotions don't lie: An audio-visual deepfake detection method using
  affective cues.
\newblock In {\em Proceedings of the 28th ACM international conference on
  multimedia}, pages 2823--2832, 2020.

\bibitem{morgado2021audio}
Pedro Morgado, Nuno Vasconcelos, and Ishan Misra.
\newblock Audio-visual instance discrimination with cross-modal agreement.
\newblock In {\em Proceedings of the IEEE/CVF Conference on Computer Vision and
  Pattern Recognition}, pages 12475--12486, 2021.

\bibitem{nagrani2018learnable}
Arsha Nagrani, Samuel Albanie, and Andrew Zisserman.
\newblock Learnable pins: Cross-modal embeddings for person identity.
\newblock In {\em Proceedings of the European Conference on Computer Vision
  (ECCV)}, pages 71--88, 2018.

\bibitem{nagrani2018seeing}
Arsha Nagrani, Samuel Albanie, and Andrew Zisserman.
\newblock Seeing voices and hearing faces: Cross-modal biometric matching.
\newblock In {\em Proceedings of the IEEE conference on computer vision and
  pattern recognition}, pages 8427--8436, 2018.

\bibitem{nguyen2019multi}
Huy~H Nguyen, Fuming Fang, Junichi Yamagishi, and Isao Echizen.
\newblock Multi-task learning for detecting and segmenting manipulated facial
  images and videos.
\newblock {\em arXiv preprint arXiv:1906.06876}, 2019.

\bibitem{niizumi2021byol}
Daisuke Niizumi, Daiki Takeuchi, Yasunori Ohishi, Noboru Harada, and Kunio
  Kashino.
\newblock Byol for audio: Self-supervised learning for general-purpose audio
  representation.
\newblock {\em arXiv preprint arXiv:2103.06695}, 2021.

\bibitem{noroozi2016unsupervised}
Mehdi Noroozi and Paolo Favaro.
\newblock Unsupervised learning of visual representations by solving jigsaw
  puzzles.
\newblock In {\em European conference on computer vision}, pages 69--84.
  Springer, 2016.

\bibitem{oord2018representation}
Aaron van~den Oord, Yazhe Li, and Oriol Vinyals.
\newblock Representation learning with contrastive predictive coding.
\newblock {\em arXiv preprint arXiv:1807.03748}, 2018.

\bibitem{park2019specaugment}
Daniel~S Park, William Chan, Yu Zhang, Chung-Cheng Chiu, Barret Zoph, Ekin~D
  Cubuk, and Quoc~V Le.
\newblock Specaugment: A simple data augmentation method for automatic speech
  recognition.
\newblock {\em arXiv preprint arXiv:1904.08779}, 2019.

\bibitem{patrick2020multi}
Mandela Patrick, Yuki~M Asano, Polina Kuznetsova, Ruth Fong, Joao~F Henriques,
  Geoffrey Zweig, and Andrea Vedaldi.
\newblock Multi-modal self-supervision from generalized data transformations.
\newblock {\em arXiv preprint arXiv:2003.04298}, 2020.

\bibitem{pinheiro2020unsupervised}
Pedro~O Pinheiro, Amjad Almahairi, Ryan~Y Benmalek, Florian Golemo, and Aaron
  Courville.
\newblock Unsupervised learning of dense visual representations.
\newblock {\em arXiv preprint arXiv:2011.05499}, 2020.

\bibitem{qian2020thinking}
Yuyang Qian, Guojun Yin, Lu Sheng, Zixuan Chen, and Jing Shao.
\newblock Thinking in frequency: Face forgery detection by mining
  frequency-aware clues.
\newblock In {\em European Conference on Computer Vision}, pages 86--103.
  Springer, 2020.

\bibitem{recasens2021broaden}
Adri{\`a} Recasens, Pauline Luc, Jean-Baptiste Alayrac, Luyu Wang, Florian
  Strub, Corentin Tallec, Mateusz Malinowski, Viorica Patraucean, Florent
  Altch{\'e}, Michal Valko, et~al.
\newblock Broaden your views for self-supervised video learning.
\newblock {\em arXiv preprint arXiv:2103.16559}, 2021.

\bibitem{rossler2019faceforensics++}
Andreas Rossler, Davide Cozzolino, Luisa Verdoliva, Christian Riess, Justus
  Thies, and Matthias Nie{\ss}ner.
\newblock Faceforensics++: Learning to detect manipulated facial images.
\newblock In {\em Proceedings of the IEEE/CVF International Conference on
  Computer Vision}, pages 1--11, 2019.

\bibitem{sabir2019recurrent}
Ekraam Sabir, Jiaxin Cheng, Ayush Jaiswal, Wael AbdAlmageed, Iacopo Masi, and
  Prem Natarajan.
\newblock Recurrent convolutional strategies for face manipulation detection in
  videos.
\newblock {\em Interfaces (GUI)}, 3(1), 2019.

\bibitem{shukla2021does}
Abhinav Shukla, Stavros Petridis, and Maja Pantic.
\newblock Does visual self-supervision improve learning of speech
  representations for emotion recognition.
\newblock {\em IEEE Transactions on Affective Computing}, 2021.

\bibitem{sun2021improving}
Zekun Sun, Yujie Han, Zeyu Hua, Na Ruan, and Weijia Jia.
\newblock Improving the efficiency and robustness of deepfakes detection
  through precise geometric features.
\newblock In {\em Proceedings of the IEEE/CVF Conference on Computer Vision and
  Pattern Recognition}, pages 3609--3618, 2021.

\bibitem{thies2019deferred}
Justus Thies, Michael Zollh{\"o}fer, and Matthias Nie{\ss}ner.
\newblock Deferred neural rendering: Image synthesis using neural textures.
\newblock {\em ACM Transactions on Graphics (TOG)}, 38(4):1--12, 2019.

\bibitem{thies2016face2face}
Justus Thies, Michael Zollhofer, Marc Stamminger, Christian Theobalt, and
  Matthias Nie{\ss}ner.
\newblock Face2face: Real-time face capture and reenactment of rgb videos.
\newblock In {\em Proceedings of the IEEE conference on computer vision and
  pattern recognition}, pages 2387--2395, 2016.

\bibitem{tian2020contrastive}
Yonglong Tian, Dilip Krishnan, and Phillip Isola.
\newblock Contrastive multiview coding.
\newblock In {\em Computer Vision--ECCV 2020: 16th European Conference,
  Glasgow, UK, August 23--28, 2020, Proceedings, Part XI 16}, pages 776--794.
  Springer, 2020.

\bibitem{tran2019video}
Du Tran, Heng Wang, Lorenzo Torresani, and Matt Feiszli.
\newblock Video classification with channel-separated convolutional networks.
\newblock In {\em Proceedings of the IEEE/CVF International Conference on
  Computer Vision}, pages 5552--5561, 2019.

\bibitem{wang2021representative}
Chengrui Wang and Weihong Deng.
\newblock Representative forgery mining for fake face detection.
\newblock In {\em Proceedings of the IEEE/CVF Conference on Computer Vision and
  Pattern Recognition}, pages 14923--14932, 2021.

\bibitem{wang2017normface}
Feng Wang, Xiang Xiang, Jian Cheng, and Alan~Loddon Yuille.
\newblock Normface: L2 hypersphere embedding for face verification.
\newblock In {\em Proceedings of the 25th ACM international conference on
  Multimedia}, pages 1041--1049, 2017.

\bibitem{wang2019fakespotter}
Run Wang, Felix Juefei-Xu, Lei Ma, Xiaofei Xie, Yihao Huang, Jian Wang, and
  Yang Liu.
\newblock Fakespotter: A simple yet robust baseline for spotting ai-synthesized
  fake faces.
\newblock {\em arXiv preprint arXiv:1909.06122}, 2019.

\bibitem{wang2020cnn}
Sheng-Yu Wang, Oliver Wang, Richard Zhang, Andrew Owens, and Alexei~A Efros.
\newblock Cnn-generated images are surprisingly easy to spot... for now.
\newblock In {\em Proceedings of the IEEE/CVF Conference on Computer Vision and
  Pattern Recognition}, pages 8695--8704, 2020.

\bibitem{wang2021dense}
Xinlong Wang, Rufeng Zhang, Chunhua Shen, Tao Kong, and Lei Li.
\newblock Dense contrastive learning for self-supervised visual pre-training.
\newblock In {\em Proceedings of the IEEE/CVF Conference on Computer Vision and
  Pattern Recognition}, pages 3024--3033, 2021.

\bibitem{wu2018unsupervised}
Zhirong Wu, Yuanjun Xiong, Stella~X Yu, and Dahua Lin.
\newblock Unsupervised feature learning via non-parametric instance
  discrimination.
\newblock In {\em Proceedings of the IEEE conference on computer vision and
  pattern recognition}, pages 3733--3742, 2018.

\bibitem{yang2019exposing}
Xin Yang, Yuezun Li, and Siwei Lyu.
\newblock Exposing deep fakes using inconsistent head poses.
\newblock In {\em ICASSP 2019-2019 IEEE International Conference on Acoustics,
  Speech and Signal Processing (ICASSP)}, pages 8261--8265. IEEE, 2019.

\bibitem{zbontar2021barlow}
Jure Zbontar, Li Jing, Ishan Misra, Yann LeCun, and St{\'e}phane Deny.
\newblock Barlow twins: Self-supervised learning via redundancy reduction.
\newblock {\em arXiv preprint arXiv:2103.03230}, 2021.

\bibitem{zeiler2014visualizing}
Matthew~D Zeiler and Rob Fergus.
\newblock Visualizing and understanding convolutional networks.
\newblock In {\em European conference on computer vision}, pages 818--833.
  Springer, 2014.

\bibitem{zhang2021deepfake}
Jian Zhang, Jiangqun Ni, and Hao Xie.
\newblock Deepfake videos detection using self-supervised decoupling network.
\newblock In {\em 2021 IEEE International Conference on Multimedia and Expo
  (ICME)}, pages 1--6. IEEE, 2021.

\bibitem{zhao2021multi}
Hanqing Zhao, Wenbo Zhou, Dongdong Chen, Tianyi Wei, Weiming Zhang, and Nenghai
  Yu.
\newblock Multi-attentional deepfake detection.
\newblock In {\em Proceedings of the IEEE/CVF Conference on Computer Vision and
  Pattern Recognition}, pages 2185--2194, 2021.

\bibitem{zhao2021learning}
Tianchen Zhao, Xiang Xu, Mingze Xu, Hui Ding, Yuanjun Xiong, and Wei Xia.
\newblock Learning self-consistency for deepfake detection.
\newblock In {\em Proceedings of the IEEE/CVF International Conference on
  Computer Vision}, pages 15023--15033, 2021.

\bibitem{zheng2021exploring}
Yinglin Zheng, Jianmin Bao, Dong Chen, Ming Zeng, and Fang Wen.
\newblock Exploring temporal coherence for more general video face forgery
  detection.
\newblock In {\em Proceedings of the IEEE/CVF International Conference on
  Computer Vision}, pages 15044--15054, 2021.

\bibitem{zhong2020random}
Zhun Zhong, Liang Zheng, Guoliang Kang, Shaozi Li, and Yi Yang.
\newblock Random erasing data augmentation.
\newblock In {\em Proceedings of the AAAI Conference on Artificial
  Intelligence}, volume~34, pages 13001--13008, 2020.

\bibitem{zhou2017two}
Peng Zhou, Xintong Han, Vlad~I Morariu, and Larry~S Davis.
\newblock Two-stream neural networks for tampered face detection.
\newblock In {\em 2017 IEEE Conference on Computer Vision and Pattern
  Recognition Workshops (CVPRW)}, pages 1831--1839. IEEE, 2017.

\bibitem{zhou2021joint}
Yipin Zhou and Ser-Nam Lim.
\newblock Joint audio-visual deepfake detection.
\newblock In {\em Proceedings of the IEEE/CVF International Conference on
  Computer Vision}, pages 14800--14809, 2021.

\bibitem{zhu2021face}
Xiangyu Zhu, Hao Wang, Hongyan Fei, Zhen Lei, and Stan~Z Li.
\newblock Face forgery detection by 3d decomposition.
\newblock In {\em Proceedings of the IEEE/CVF Conference on Computer Vision and
  Pattern Recognition}, pages 2929--2939, 2021.

\end{thebibliography}
}

\newpage
\appendix

\section{More Experiments}
\subsection{In-distribution performance}
Although our approach has been developed for cross-manipulation generalisation and robustness, for completeness we present results for in-distribution performance in Table \ref{table:ff++}. For each compression level (raw, c23, c40), we train on the training set and show results on the corresponding test set. We are on par with the state-of-the-art in the no/low compression regime, while outperforming the other methods on the more compressed data.

\begin{table}
\begin{center}
\resizebox{\linewidth}{!}{
\begin{tabular}{l c c c c c c}\toprule
\multirow{2}{*}{Method} & \multicolumn{3}{c}{Accuracy (\%)} & \multicolumn{3}{c}{AUC (\%)} \\ \cmidrule(lr){2-4} \cmidrule(lr){5-7}
 & Raw & c23 & c40 & Raw & c23 & c40 \\ \midrule 
Xception \cite{rossler2019faceforensics++} & 99.0 & 97.0 & 89.0 & 99.8 & 99.3 & 92.0 \\
CNN-aug \cite{wang2020cnn} & 98.7 & 96.9 & 81.9 & 99.8 & 99.1 & 86.9  \\
Patch-based \cite{chai2020makes} & \textbf{99.3} & 92.6 & 79.1 & \textbf{99.9} & 97.2 & 78.3 \\
Two-branch \cite{masi2020two} & --- & --- & --- & --- & 99.1 & 91.1 \\
Face X-ray \cite{li2020face} & 99.1 & 78.4 & 34.2 & 99.8 & 97.8 & 77.3 \\
CNN-GRU \cite{sabir2019recurrent} & 98.6 & 97.0 & 90.1 & \textbf{99.9} & 99.3 & 92.2 \\
LipForensics \cite{haliassos2021lips} & 98.9 & 98.8 & 94.2 & \textbf{99.9} & 99.7 & 98.1 \\
FTCN \cite{zheng2021exploring} & --- & \textbf{99.1} & --- & --- & \textbf{99.8} & 98.3 \\ \midrule
RealForensics (ours) & \textbf{99.3} & \textbf{99.1} & \textbf{96.1} & \textbf{99.9} & \textbf{99.8} & \textbf{99.5} \\ \bottomrule
\end{tabular}
}
\end{center}
\caption{\textbf{In-distribution performance.} Accuracy and AUC scores on the test set of FaceForensics++ (FF++) after training on FF++. We repeat experiments for the dataset's three compression types: raw (no compression), c23 (mild compression), and c40 (strong compression. Best results are in \textbf{bold}.}
\label{table:ff++}
\end{table}

\subsection{Generalisation to ForgeryNet}
In Table \ref{table:forgerynet}, we provide results on generalisation performance to the newly-released ForgeryNet dataset \cite{he2021forgerynet}. We compare our model with the publicly-available LipForensics and FTCN models (all trained on FF++). RealForensics significantly outperforms both. 

\begin{table}
\begin{center}
\begin{tabular}{l | c c c}\toprule
 & Ours & LipForensics \cite{haliassos2021lips} & FTCN \cite{zheng2021exploring} \\ \midrule
ForgeryNet & \textbf{71.8} & 66.7 & 57.3 \\
 \bottomrule
\end{tabular}
\end{center}
\caption{\textbf{Generalisation to ForgeryNet.} AUC scores (\%) on the val set of ForgeryNet after training on FF++. Best results are in \textbf{bold}.}
\label{table:forgerynet}
\end{table}

\subsection{Detailed analysis of robustness} \label{sec:detail_robustness}
\begin{figure*}
\centering
  \centerline{\includegraphics[width=\linewidth]{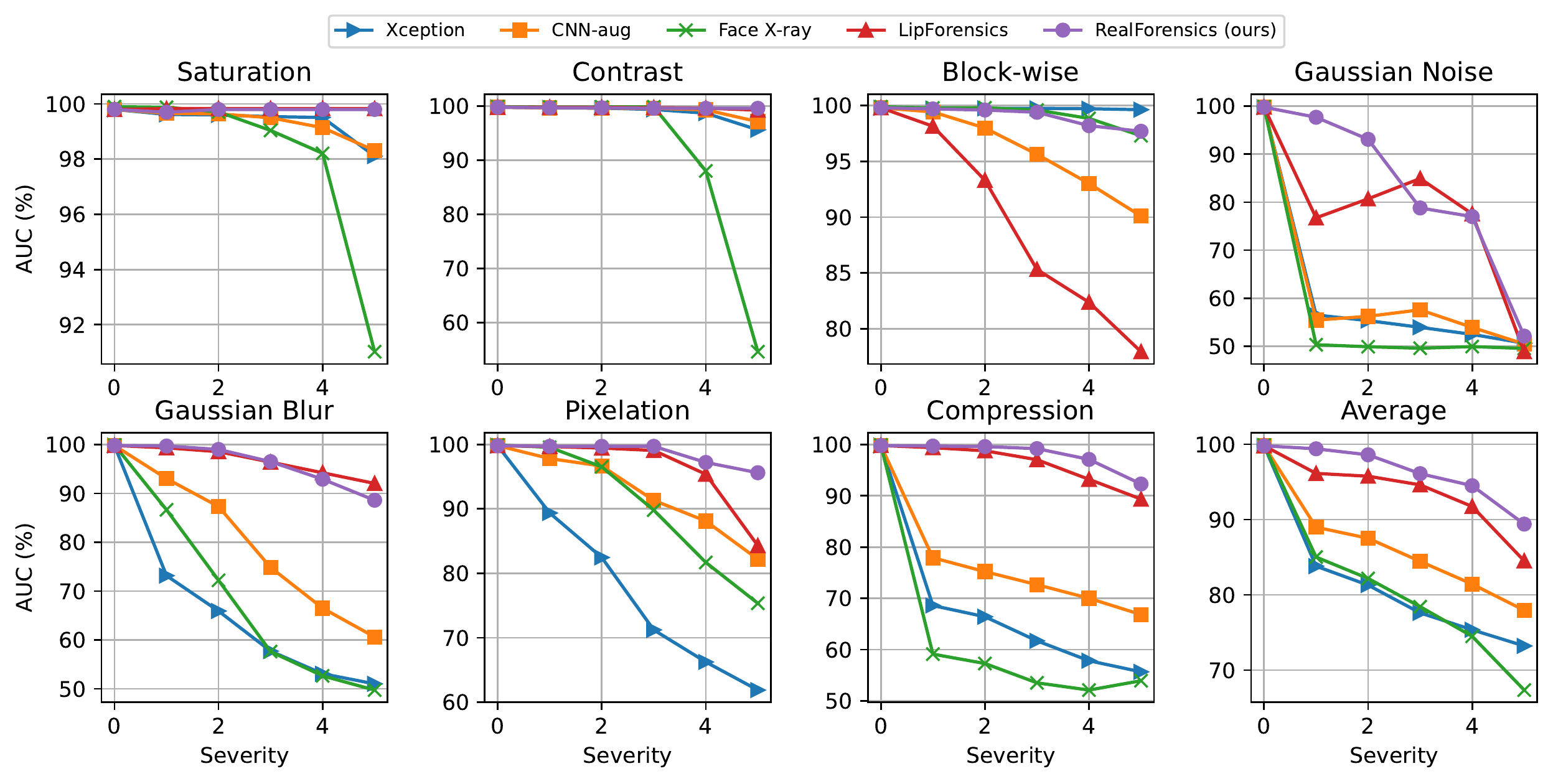}}
 \caption{\textbf{Robustness to unseen perturbations.} AUC scores (\%) on FaceForensics++ samples which have been corrupted by various unseen perturbations of varying severity. We also present the average scores across all perturbations. All methods were trained on FF++ without these corruptions. To avoid visual clutter in the plots, we show results for five representative methods. For more results, see \cite{haliassos2021lips} and \cite{zheng2021exploring}.}
 \label{fig:corruption_severity}
\end{figure*}

Following \cite{haliassos2021lips}, we present more detailed results on robustness by plotting AUC as a function of corruption severity (see Figure \ref{fig:corruption_severity}). On average, RealForensics deteriorates less abruptly as severity increases than other methods, with especially noteworthy results on video compression, which is ubiquitous on social media. We also highlight our significantly higher results over LipForensics on block-wise distortions (\textit{i.e.}, occlusions), which are likely influenced by our method's use of the whole face rather than solely the mouth. For example, in some cases the mouth may be occluded while other parts of the face are not.

\begin{figure}
\centering
  \centerline{\includegraphics[width=\linewidth]{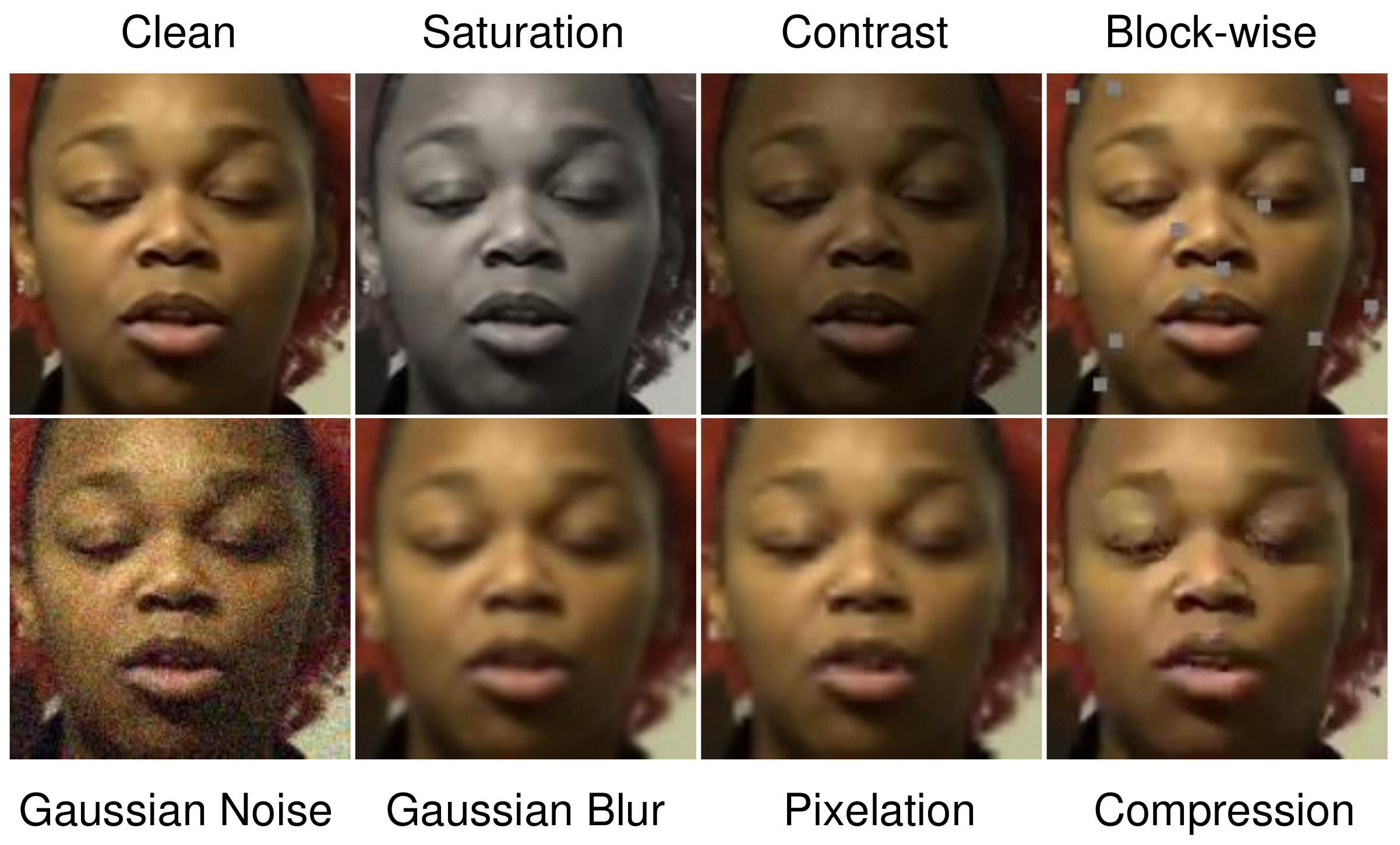}}
 \caption{\textbf{Examples of corruptions.} A clean frame from a real FaceForensics++ video along with the same frame but corrupted with various perturbations. For more information on this set of corruptions, see \cite{jiang2020deeperforensics}.}
 \label{fig:corruption_examples}
\end{figure}

\subsection{More ablations} \label{sec:more_ablations}
\begin{description}[wide,itemindent=\labelsep]
\item[Full face versus mouth.] In the main text, we argue that focusing only on the mouth region, like LipForensics \cite{haliassos2021lips}, may be suboptimal for performance. We validate this by training (for both stages 1 and 2) on mouth crops and comparing the performance with the default setting. As shown in Table \ref{table:fullface_vs_mouth}, our method consistently benefits from using the full face rather than the mouth, which was not observed for LipForensics \cite{haliassossupplementary}. This may be due to the cross-modal prediction task being more general than lipreading. For example, the video network is encouraged to retain information about the eyes to better model expression (which correlates with audio); on the other hand, a model trained to perform lipreading may focus predominantly on the mouth region.

\begin{table}
\begin{center}
\begin{tabular}{l c c c}\toprule
\multirow{2}{*}{Crop} & \multicolumn{2}{c}{Acc (\%)} & AUC (\%) \\  
 \cmidrule(lr){2-3} \cmidrule(lr){4-4}
 & FSh & DFo & FS \\ \midrule
\rowcolor{light-gray}
Full face & \textbf{97.1} & \textbf{97.1} & \textbf{97.1} \\
Mouth & 95.5 & 95.0 & 88.9 \\ \bottomrule
\end{tabular}
\end{center}
\caption{\textbf{Full face versus mouth.} Accuracy and AUC scores when training on full faces and mouth crops. We test on FaceShifter (FSh) and DeeperForensics (DFo) after training on FaceForensis++ (FF++). We also test on FaceSwap (FS) after training on the remaining three FF++ types. Best results are in \textbf{bold}. Default setting is \colorbox{light-gray}{highlighted}.}
\label{table:fullface_vs_mouth}
\end{table}

\item[Effect of clip size.] Table \ref{table:clip_size} shows the effect on generalisation when changing the video clip size (default is 25 frames per clip). We observe that generalisation improves with clip size, up to a point.

\begin{table}
\begin{center}
\footnotesize
\begin{tabular}{l | c c c c c c}\toprule
Clip size (\# frames) & 5 & 10 & 15 & 20 & 25 & 30 \\ \midrule
DeeperForensics & 88.2 & 95.0 & 96.1 & 96.4 & 97.1 & \textbf{97.4} \\
FaceShifter & 87.9 & 93.4 & 95.4 & 95.7 & \textbf{97.1} & 96.7 \\
 \bottomrule
\end{tabular}
\end{center}
\caption{\textbf{Effect of clip size.} Accuracy (\%) as a function of the clip size. We test on FaceShifter and DeeperForensics after training on FaceForensis++. Best results are in \textbf{bold}.}
\label{table:clip_size}
\end{table}

\item[Different backbone.] Our default video backbone is a CSN network \cite{tran2019video}. In Table \ref{table:backbone} we also show generalisation results for ResNet+MS-TCN \cite{martinez2020lipreading}, used in \cite{haliassos2021lips}. We significantly outperform LipForensics with the same backbone and auxiliary dataset (compare with Table 3 in the main text), without requiring any auxiliary labels.

\begin{table}
\begin{center}
\begin{tabular}{l c c}\toprule
Backbone & FSh & DFo \\ \midrule
\rowcolor{light-gray}
CSN & \textbf{97.1} & \textbf{97.1} \\
ResNet+MS-TCN & 94.0 & 95.7 \\ \bottomrule
\end{tabular}
\end{center}
\caption{\textbf{Backbones.} Accuracy scores (\%) on FaceShifter (FSh) and DeeperForensics (DFo) after training on FaceForensics++. We show results for two different backbones. Best results are in \textbf{bold}. Default setting is \colorbox{light-gray}{highlighted}.}
\label{table:backbone}
\end{table}

\item[Projector and predictor.] We propose in the main text to use a single linear layer as our projector and a shallow transformer as the predictor. In Table \ref{table:projector_and_predictor}, we show generalisation results when using different types of projectors/predictors. Since we output dense representations, the linear layers in the MLPs can be thought of as convolutional layers with kernel size 1. We use a learning rate of $3\times 10^{-4}$ when employing MLP predictors, as we found it to perform best in that setting. 

Notably, we observe that using a transformer improves results over the MLP variant. This suggests that allowing the predictor to model temporal dynamics can benefit representation learning for our task. Further, in Table \ref{table:transformer_layers} we show results for a 1-block and a 2-block transformer predictor. We find that the 1-block variant performs slightly better.

\begin{table}
\begin{center}
\begin{tabular}{c c c c c}\toprule
\multicolumn{2}{c}{Settings} & \multicolumn{2}{c}{Accuracy (\%)} \\ \cmidrule(lr){1-2} \cmidrule(lr){3-4}
Projector & Predictor & FSh & DFo  \\ \midrule
Linear & MLP & 91.8 & 92.9 \\
\rowcolor{light-gray}
Linear & Transformer & \textbf{97.1} & 97.1 \\
MLP & MLP & 91.1 & 92.5 \\
MLP & Transformer & 96.1 & \textbf{97.5} \\ \bottomrule 
\end{tabular}
\end{center}
\caption{\textbf{Projector and predictor.} We test different types of projectors and predictors for the representation learning stage of our method (stage 1), and see how generalisation to FaceShifter (FSh) and DeeperForensics (DFo) is affected after training on FaceForensics++. Refer to subsection ``Projector and predictor'' for a discussion. Best results are in \textbf{bold}. Default setting is \colorbox{light-gray}{highlighted}.}
\label{table:projector_and_predictor}
\end{table}

\begin{table}
\begin{center}
\begin{tabular}{c c c}\toprule
\# blocks & FSh & DFo \\ \midrule
\rowcolor{light-gray}
1 & \textbf{97.1} & \textbf{97.1} \\
2 & 96.8 & 96.4 \\ \bottomrule
\end{tabular}
\end{center}
\caption{\textbf{Number of transformer blocks.} Accuracy scores (\%) on FaceShifter (FSh) and DeeperForensics (DFo) after training on FaceForensics++. We show results for a 1-block and a 2-block transformer predictor. Best results are in \textbf{bold}. Default setting is \colorbox{light-gray}{highlighted}.}
\label{table:transformer_layers}
\end{table}

\item[Different contrastive baselines.] As mentioned in the main text, self-supervised methods that aim to learn representations for lipreading tend to contrast samples from the same video to achieve invariance to identity \cite{chung2019perfect, chung2016out, afouras2020self}. Here, instead of our proposed non-contrastive approach, we apply the strategy of the audiovisual method Perfect Match \cite{chung2019perfect} for stage 1 of our method. For fair comparison, we use the same backbones as for RealForensics. We follow the instructions from the paper for implementation. In particular, the inputs to the video and audio backbones are 5-frame video clips and 20-frame log mel spectrograms. Each network yields a single feature (via a temporal pooling layer). Then, for a single video feature, a contrastive loss is employed to match it to its aligned audio feature while repelling misaligned ones from the same video. We found that symmetrising this loss by additionally adding the loss corresponding to the reversal of the roles of the video and audio features yielded improvements; we refer to this variant as Perfect Match++. The results in Table \ref{table:pmatch} suggest that our proposed method, which does not target identity invariance, is better suited for forgery detection. 

\begin{table}
\begin{center}
\begin{tabular}{l c c}\toprule
Method & FSh & DFo \\ \midrule
PMatch & 91.4 & 87.9 \\
PMatch++ & 91.8 & 90.2 \\ \midrule
RealForensics (ours) & \textbf{97.1} & \textbf{97.1} \\ \bottomrule
\end{tabular}
\end{center}
\caption{\textbf{Different contrastive baselines.} Accuracy scores (\%) on FaceShifter (FSh) and DeeperForensics (DFo) after training on FaceForensics++. We show results by employing the learning strategy of Perfect Match \cite{chung2019perfect} (PMatch) for stage 1 of our method. We also use a symmetrised version of Perfect Match, which we call PMatch++. Best results are in \textbf{bold}.}
\label{table:pmatch}
\end{table}

\item[Visual-only representation learning.] Although it is natural to use the correspondence between the visual and auditory modalities to capture information related to facial behaviour and appearance, we present here some preliminary results on using only the visual modality in the representation learning stage. To this end, we extend BYOL to the video setting by using a single student-teacher pair. As is the case for the cross-modal task, the network outputs temporally dense representations, and we use a transformer for the predictor. We apply the augmentations proposed in \cite{grill2020bootstrap} to each frame, consistently across the whole video. The results in Table \ref{table:visual_only} indicate that our proposed cross-modal task strongly benefits generalisation, likely because audiovisual correspondence provides
a richer signal for encoding natural facial movements and
expressions. We leave for future work the investigation of more effective video augmentations that could further improve the visual-only baseline.

\begin{table}
\begin{center}
\begin{tabular}{l c c}\toprule
Type & FSh & DFo \\ \midrule
Visual & 92.9 & 89.7 \\ 
\rowcolor{light-gray}
Audiovisual & \textbf{97.1} & \textbf{97.1} \\ \bottomrule
\end{tabular}
\end{center}
\caption{\textbf{Visual versus audiovisual representation learning.} Accuracy scores (\%) on FaceShifter (FSh) and DeeperForensics (DFo) after training on FaceForensics++. We compare visual-only with audiovisual representation learning (using BYOL-style training) for stage 1 of our method. Best results are in \textbf{bold}. Default setting is \colorbox{light-gray}{highlighted}.}
\label{table:visual_only}
\end{table}

\end{description}

\section{Further Implementation Details} \label{sec:more_implementation_details}
\subsection{Preprocessing}
We use RetinaFace \cite{deng2020retinaface}\footnote{\url{https://github.com/biubug6/Pytorch_Retinaface}} for face detection and a 2-D FAN network \cite{bulat2017far}\footnote{\url{https://github.com/1adrianb/face-alignment}} to extract 68 facial landmarks. For each frame, we take the mean landmarks around a 12-frame window to reduce motion jitter and then affine warp to LRW's mean face based on eight stable points.

\subsection{Dataset details} \label{sec:datasets}

We provide further details on the used datasets. The licenses of all datasets permit their use for research purposes.

\begin{description}[wide,itemindent=\labelsep]
\item[FaceForensics++ \cite{rossler2019faceforensics++} (FF++).] We use the dataset from the official webpage\footnote{\url{https://github.com/ondyari/FaceForensics}}. We use the provided train/validation/test splits, which include 720 training, 140 validation, and 140 test videos, respectively.
\item[FaceShifter \cite{li2020advancing}.] We use the dataset (at compression c23) from the FF++ webpage. Its real videos come from FF++. Note that we do not treat FaceShifter as part of FF++, consistent with the original paper \cite{rossler2019faceforensics++}.

\item[DeeperForensics \cite{jiang2020deeperforensics}.] We use the dataset from the official webpage\footnote{\url{https://github.com/EndlessSora/DeeperForensics-1.0/tree/master/perturbation}}. Its real videos also come from FF++ c23.

\item[CelebFD-v2 \cite{li2020celeb}.] We use the dataset from the official webpage\footnote{\url{https://github.com/yuezunli/celeb-deepfakeforensics}}.

\item[DFDC \cite{dolhansky2020deepfake}.] We use a subset of the dataset from the official webpage\footnote{\url{https://ai.facebook.com/datasets/dfdc}}. This subset was used in \cite{haliassos2021lips} and features single-subject videos for which the face and landmark detectors did not fail (since many videos have been subjected to extreme perturbations).

\end{description}

\subsection{Architecture and training details} 
\label{sec:arch_train_details}

\begin{description}[wide,itemindent=\labelsep]
\item[Supervised loss details.] As described in Section 3.2 of the main text, we use a cosine classifier for our supervised head and also employ logit adjustment \cite{menon2020long} to address data imbalance. Given the (average-pooled) output $e$ of the backbone network and the weight vector $w$ of the supervised head's linear layer, the normalised score of a sample's ``fakeness'' \textit{during training} is given as 
\begin{align}
    p = \frac{1}{1 + e^{-\left(s\frac{w\cdot e}{\|w\|_2\|e\|_2} + \log\frac{\pi}{1 - \pi}\right)}},
\end{align}
where $s=64$ scales the cosine similarity, as in e.g., \cite{wang2017normface}, and $\pi$ is the prior probability of a sample being fake, as described in \cite{menon2020long}. We set $\pi$ to be the ratio of fake samples to the batch size. We found using cosine similarity (\textit{i.e.}, normalising the feature and weight vectors) yielded slight improvements; the ablation on logit adjustment is given in Table 5 of the main text. The supervised loss $\mathcal{L}_s(\mathcal{D}; \theta_b, \theta_s)$, introduced in Section 3.2 of the main text, is simply the standard binary cross entropy acting on these scores.

\item[Random masking.] We apply random erasing to video frames with probability 0.5, scale of $(0.02, 0.33)$, and ratio of $(0.3, 3.3)$. Moreover, we randomly erase a random number of video frames, ranging from 0 to 12, a random number of audio frames, ranging from 0 to 48, and a random number of mel filters, ranging from 0 to 27. This is applied with probability 0.5.

\item[Backbones.] Our video backbone is a modified Channel-Separated Convolutional Network (CSN) \cite{tran2019video}, chosen for its high accuracy in video action recognition \cite{tran2019video} in conjunction with its relatively low parameter count. Unlike the original architecture, we set the temporal strides to 1 for all layers, thus preserving the temporal dimension. See Table \ref{table:csn_arch} for more information.

Our audio backbone is a ResNet18 \cite{he2016deep}. We modify the temporal strides to match the output size of the video backbone. In particular, the stem subsamples the temporal dimension by 4, after which no further temporal subsampling is performed. See Table \ref{table:rn_arch} for more information.

\item[Details on MLPs used in ablations.] In the ablations where we use MLPs for the projector and/or predictor, we follow the design proposed in \cite{chen2021exploring}, as we found it to perform well. Thus, the projector MLP has 3 layers with hidden dimension 2048, and each layer is followed by batch normalisation (BN); the output layer has no ReLU activation. The predictor MLP has 2 layers with hidden dimension 512 and output dimension 2048, and the output layer has no BN nor ReLU.

\item[Further details on contrastive experiments.]
We provide more details on the experiments with contrastive learning given in Table 6 of the main text. For dense representation learning, the output of the network consists of 25 embeddings (one for each video frame); we select a random embedding to add to the queue of negative samples. We also use shuffling batch normalisation to prevent the network from cheating on the pretext task \cite{he2020momentum}.

\end{description}

\begin{table}[tb]
\begin{center}
\resizebox{\linewidth}{!}{
\begin{tabular}{c | c | c}
stage & filters & output size  \\ \toprule
conv\textsubscript{1} & $3\times 7\times 7$, stride $1\times 2\times 2$ & $25\times 56\times 56$ \\ \midrule
pool\textsubscript{1} & max, $1\times 3\times 3$, stride $1\times 2\times 2$ & $25\times 28\times 28$ \\ \midrule
res\textsubscript{1} & $\begin{bmatrix} 1\times 1\times 1, 256 \\ 3\times 3\times 3, 64 \\ 1\times 1\times 1, 256 \end{bmatrix} \times 3$ & $25\times 28\times 28$ \\ \midrule
res\textsubscript{2} & $\begin{bmatrix} 1\times 1\times 1, 512 \\ 3\times 3\times 3, 128 \\ 1\times 1\times 1, 512 \end{bmatrix} \times 4$ & $25\times 14\times 14$ \\ \midrule
res\textsubscript{3} & $\begin{bmatrix} 1\times 1\times 1, 1024 \\ 3\times 3\times 3, 256 \\ 1\times 1\times 1, 1024 \end{bmatrix} \times 23$ & $25\times 7\times 7$ \\ \midrule
res\textsubscript{4} & $\begin{bmatrix} 1\times 1\times 1, 2048 \\ 3\times 3\times 3, 512 \\ 1\times 1\times 1, 2048 \end{bmatrix} \times 3$ & $25\times 4\times 4$ \\ \midrule
pool\textsubscript{2} & global spatial average pool & $25\times 1\times 1$ \\ \midrule
\end{tabular}
}
\end{center}
\caption{\textbf{Video backbone architecture.} The architecture of the modified CSN \cite{tran2019video} network that we employ for the video backbone. The layers in the bottleneck blocks, shown in brackets, use \textit{depthwise convolutions}. Next to the brackets we give the number of times the blocks are repeated in each stage. The output size is of the form $T\times H\times W$, where $T$ denotes time, \textit{H} height, and $W$ width. Note that differently from the original architecture \cite{tran2019video}, we do not subsample the temporal dimension at any stage and also only use spatial pooling at the end, rather than spatio-temporal, since we employ dense learning.}
\label{table:csn_arch}
\end{table}

\begin{table}[tb]
\begin{center}
\begin{tabular}{c | c | c}
stage & filters & output size  \\ \toprule
conv\textsubscript{1} & $7\times 7$, stride $2\times 2$ & $50\times 40$ \\ \midrule
pool\textsubscript{1} & max, $3\times 3$, stride $2\times 2$ & $25\times 20$ \\ \midrule
res\textsubscript{1} & $\begin{bmatrix} 3\times 3, 64 \\ 3\times 3, 64 \end{bmatrix} \times 2$ & $25\times 20$ \\ \midrule
res\textsubscript{2} & $\begin{bmatrix} 3\times 3, 128 \\ 3\times 3, 128 \end{bmatrix} \times 2$ & $25\times 10$ \\ \midrule
res\textsubscript{3} & $\begin{bmatrix} 3\times 3, 256 \\ 3\times 3, 256 \end{bmatrix} \times 2$ & $25\times 5$ \\ \midrule
res\textsubscript{4} & $\begin{bmatrix} 3\times 3, 512 \\ 3\times 3, 512 \end{bmatrix} \times 2$ & $25\times 3$ \\ \midrule
pool\textsubscript{2} & global frequency average pool & $25\times 1$ \\ \midrule
\end{tabular}
\end{center}
\caption{\textbf{Audio backbone architecture.} The architecture of our modified ResNet18 \cite{he2016deep} network that we employ for the audio backbone. The layers in a residual blocks are in brackets, next to which we give the number of times the blocks are repeated in each stage. The output size is of the form $T\times F$, where $T$ denotes time and $F$ mel filters. Note that differently from the original architecture \cite{he2016deep}, we do not subsample the temporal dimension at any stage and also only use mel frequency pooling at the end, since we employ dense learning.}
\label{table:rn_arch}
\end{table}

\section{Visualisation}
\begin{figure}
\centering
  \centerline{\includegraphics[width=\linewidth]{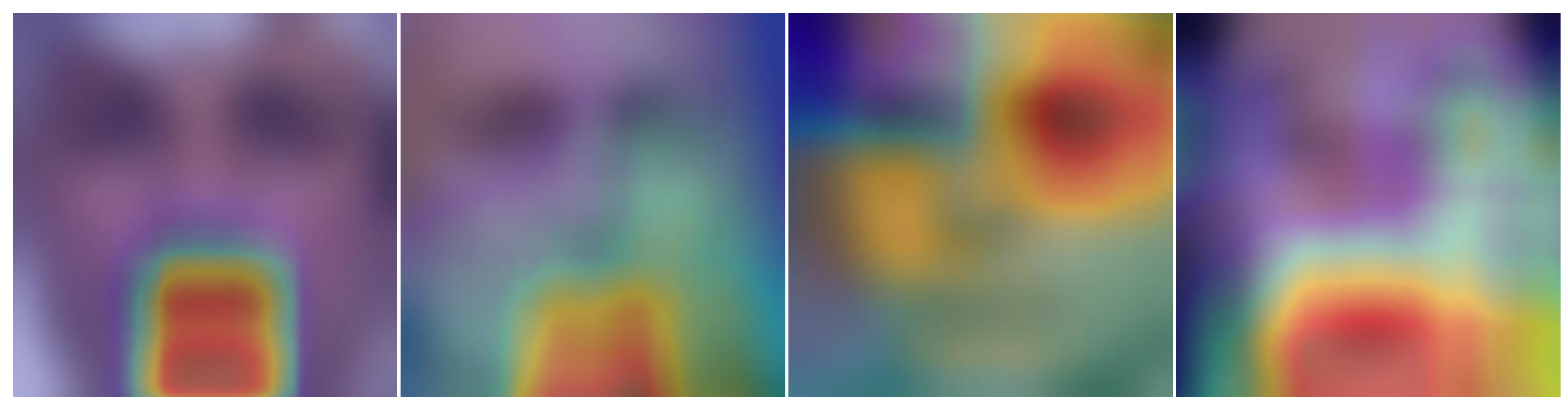}}
 \caption{\textbf{Occlusion sensitivity analysis.} Occlusion sensitivity examples for FaceForensics++ types. The faces have been blurred to preserve anonymity.}
 \label{fig:occlusion_sensitivity}
\end{figure}

We use occlusion sensitivity analysis \cite{zeiler2014visualizing} for visualisation, as in \cite{haliassos2021lips}. We systematically occlude, in a sliding-window fashion, parts of the video via random erasing of size $40\times 40\times T$ (where $T$ is the number of frames). We record for each occluded pixel the effect that the occlusion has on the model predictions. A heatmap is produced by averaging the output probabilities for each pixel. After normalisation, we overlay the heatmap on the first video frame. We show examples for FaceForensics++ in Figure \ref{fig:occlusion_sensitivity}. We see that for NeuralTextures and Face2Face (first two examples), which modify expressions, our network usually focuses on the mouth region. On the face-swapping types, we observe that sometimes the network focuses on the mouth and sometimes on other facial regions.

\end{document}